\documentclass[runningheads]{llncs}

\usepackage{eccv}

\usepackage{eccvabbrv}

\usepackage{graphicx}
\usepackage{booktabs}

\usepackage[accsupp]{axessibility}  

\usepackage{hyperref}

\usepackage{orcidlink}

\definecolor{purple}{rgb}{0.65,0,0.65}
\definecolor{dark_green}{rgb}{0, 0.5, 0}
\definecolor{blueish}{rgb}{0.0, 0.3, .6}
\definecolor{LightCyan}{rgb}{0.88,0.95,1}
\definecolor{tabhighlight}{HTML}{e5e5e5}

\newcommand{\custompar}[1]{\paragraph{\textbf{#1}}}

\usepackage{array}
\newcolumntype{H}{>{\setbox0=\hbox\bgroup}c<{\egroup}@{}}

\begin{document}

\title{Neuromorphic Drone Detection: an Event-RGB Multimodal Approach}
\titlerunning{Neuromorphic Drone Detection: an Event-RGB Multimodal Approach}

\author{Gabriele Magrini\inst{1}\orcidlink{0009-0004-1341-5931} \and
Federico Becattini\inst{2}\orcidlink{0000-0003-2537-2700} \and
Pietro Pala\inst{1}\orcidlink{0000-0001-5670-3774}\and
Alberto Del Bimbo\inst{1}\orcidlink{0000-0002-1052-8322} \and
Antonio Porta\inst{3}\orcidlink{}
}

\authorrunning{G.~Magrini et al.}

\institute{University of Florence, Italy
\email{name.surname@unifi.it} \and
University of Siena, Italy
\email{name.surname@unisi.it} \and
Leonardo S.p.A.
\email{name.surname@leonardo.com}}

\maketitle

\begin{abstract}
  In recent years, drone detection has quickly become a subject of extreme interest: the potential for fast-moving objects of contained dimensions to be used for malicious intents or even terrorist attacks has posed attention to the necessity for precise and resilient systems for detecting and identifying such elements.
  While extensive literature and works exist on object detection based on RGB data, it is also critical to recognize the limits of such modality when applied to UAVs detection. Detecting drones indeed poses several challenges such as fast-moving objects and scenes with a high dynamic range or, even worse, scarce illumination levels. 
  Neuromorphic cameras, on the other hand, can retain precise and rich spatio-temporal information in situations that are challenging for RGB cameras. They are resilient to both high-speed moving objects and scarce illumination settings, while prone to suffer a rapid loss of information when the objects in the scene are static. In this context, we present a novel model for integrating both domains together, leveraging multimodal data to take advantage of the best of both worlds. To this end, we also release \textbf{NeRDD} (\textbf{Ne}uromorphic-\textbf{R}GB \textbf{D}rone \textbf{D}etection), a novel spatio-temporally synchronized Event-RGB Drone detection dataset of more than 3.5 hours of multimodal annotated recordings. 
\end{abstract}

\section{Introduction}
\label{sec:intro}

Drones are versatile devices with a wide range of applications, including photography, videography, agriculture, search and rescue operations, environmental monitoring, infrastructure inspection, and more. However, concerns about privacy have also been raised due to the potential for drones to capture images, videos and audio of individuals without their consent~\cite{altawy2016security}.
In response to these concerns, many countries have enacted regulations governing the use of drones and addressing privacy issues. These regulations often include guidelines for where drones can be flown, how high they can fly, and restrictions on capturing images or videos of private property or individuals without permission\footnote{\url{https://www.easa.europa.eu/en/the-agency/faqs/drones-uas}}.

Advancements in technology are being developed to help mitigate potential privacy risks associated with drone use \cite{Seidaliyeva-2024, Yousaf-2022}. 
For instance, geofencing technology creates virtual boundaries using GPS or RFID. Drones equipped with geofencing capabilities can be programmed to avoid restricted areas, such as private properties, government buildings, and sensitive locations. This ensures drones do not inadvertently capture images or data from these areas, thus protecting privacy.
Additionally, to defend from both unintentional abuse and deliberate attack, a variety of methods and technologies have been proposed to detect drones without relying on their deliberate cooperation. 
\emph{Radio Frequency (RF) Analysis} is widely used for drone detection and operates by detecting the radio signals used for drone communication and control~\cite{al2020drone}. However, this technology is of limited efficacy when the drone is equipped with processing modules enabling it to operate autonomously. 
\emph{High-resolution radar} systems can detect and track drones by sending out signals and analyzing the reflections from objects in the sky~\cite{nuss2017mimo}. However, traditional radar technology can struggle to detect increasingly miniaturized commercial drones, many of which have the size of a bird. Even if the radar system can detect very small objects it could not be able to distinguish a small drone from a bird.
\emph{Acoustic Sensors} use microphones to detect the unique sound signatures produced by drones' propellers and engines~\cite{svanstrom2021real, svanstrom2021real}. However, this technology does not work as well in noisy environments and it also has a very short operative range.
\emph{Optical Sensors} including RGB~\cite{yang2022drone} and thermal imaging cameras~\cite{svanstrom2021real}, can visually detect drones~\cite{elsayed2021review}. Thermal cameras are particularly effective for drone detection as they can capture the heat signatures of drone propellers. However, only cooled thermal cameras are capable of capturing such signatures at a distance, which results in bulky and rather expensive devices. 
Furthermore, in strong sunlight, the ambient temperature of parts of the environment can rise significantly, causing non-target objects (like buildings, roads, and vegetation) to emit infrared radiation. This causes clutter and reduces the contrast between the temperature of the heated propellers and their surroundings, making it harder to detect them.
Also, the use of RGB cameras for drone detection is particularly challenging when the drone is observed under a cluttered background. 
In such conditions, even state-of-the-art detection models, such as recent versions of the YOLO network, fail to adequately address the detection task \cite{Mistry-2023}. 

Recently, neuromorphic cameras, also referred to as event-based cameras or dynamic vision sensors (DVS), have been introduced to advance imaging technology in scenarios involving fast-moving objects and varying illumination conditions \cite{Gallego-2022}. These cameras operate by detecting changes in the scene at the level of individual pixels asynchronously. This allows them to capture events with very high temporal resolution, often in the microsecond range, which is ideal for fast-moving objects. 
In addition, these cameras can handle a wide range of illumination conditions, from very low light to extremely bright environments, without suffering from issues like saturation or loss of detail. This high dynamic range is beneficial for outdoor applications or environments with variable lighting.
Finally, since neuromorphic cameras only record changes in the scene rather than capturing full frames at regular intervals, they produce less redundant data compared to traditional frame-based cameras, resulting in lower data rates and more efficient processing.
This high temporal resolution and ability to handle erratic movements make neuromorphic cameras well-suited for tasks such as detection and tracking of drones, even those with rapid and unpredictable flight trajectories.
At the same time, traditional optical sensors could offer an advantage over event cameras, if the drone is stationary. Motivated by these reasons, we propose a multimodal approach that leverages both neuromorphic and RGB data.
In this paper, we study in detail how the two modalities can be fused into a single end-to-end trained detection model, analyzing the impact of different strategies ranging from simple pooling to more complex attention-based solutions.
The main contributions of the paper are the following:
\begin{itemize}
    \item We present a multimodal architecture to detect drones based on neuromorphic and RGB data. To the best of our knowledge, we are the first to combine the two modalities and report on the accuracy of different fusion strategies for drone detection.
    \item We release \textbf{NeRDD}\footnote{\url{https://github.com/MagriniGabriele/NeRDD}}, a \textbf{Ne}uromorphic-\textbf{R}GB \textbf{D}rone \textbf{D}ataset, comprising more than 3.5 hours of spatio-temporally synchronized event and RGB data, manually annotated with drone bounding boxes. To the best of our knowledge, NeRDD is seven fold larger than the only other neuromorphic drone dataset existing in the literature.
    \item We demonstrate the effectiveness of drone detection with event cameras, compared to the less effective RGB counterpart. Our experiments show that by combining the two modalities, we can further improve the detection rate.
\end{itemize}

\section{Related Work}
\label{sec:prev-work}

\custompar{Neuromorphic Object Detection}
Approaches to object detection with neuromorphic cameras can be broadly grouped into two classes, depending on whether the stream of events is processed by preserving the spatio-temporal sparsity of the events or by first converting the stream to dense, pseudo-frame representations.
Among the former approach, several methods have been proposed for object detection using Spiking Neural Networks \cite{Nan-2019, Li-2019, Liu-2020, Yao-2021, Cordone-2022, Zhang-2022}, biologically inspired networks composed of neurons that communicate using discrete and asynchronous spikes.
Such an approach has been used in several fields, such as automotive for vehicle and pedestrian detection \cite{Cordone-2022}. To improve detection rates, solutions like temporal-wise attention have been studied \cite{Yao-2021}, as well as multi-camera processing involving two neuromorphic sensors \cite{Li-2019}.
Zhang et al. \cite{Zhang-2022} adopted a spiking transformer network, STNet, to detect and track objects leveraging both spatial and temporal information.

Other approaches have recently gained popularity, leveraging different types of architectures such as Graph Neural Networks \cite{Bi-2019} or convolutional detectors borrowed from the RGB literature \cite{becattini2022understanding}.
In \cite{Schaefer-2022}, a generalized GNN architecture is proposed
where events are represented as nodes within the graph and edges are formed between neighboring events in the spatio-temporal domain. Each new event yields a local change to the activation of graph nodes whose outputs propagate asynchronously across network layers. 
Transformer-based architectures have also started to make their appearance in the neuromorphic domain \cite{Gehrig-2023, Pei-2024}. In \cite{Gehrig-2023}, a novel backbone for object detection in event streams is proposed, which relies on a novel Recurrent Vision Transformers (RVTs) architecture. 
A transformer architecture combined with a graph neural network to represent the stream of events is also exploited in \cite{Pei-2024} to perform action classification.
As an alternative strategy for representing events, a consolidated line of research relies on the aggregation of events across time to convert acquired data into a sequence of pseudo-frames each one collecting the events occurring in a time interval of predefined length. 
This process converts the data stream into a dense representation in the spatio-temporal domain, enabling to leverage well established convolutional backbone architectures for data analysis \cite{nguyen2019real, berlincioni2023neuromorphic, innocenti2021temporal, mueggler2017fast}.
We rely on this strategy as it represents a good tradeoff between simplicity and effectiveness of the approach.
Detection and classification of objects in event streams is also addressed in the approaches reported in \cite{Damien-2019, Ramesh-2020, Chen-2019} that rely on the aggregation of events across time to construct a matrix-like representation suitable for being processed by 2D filters. 
In \cite{Ramesh-2020}, this filtered event-count patch is then projected on-to a lower-dimensional subspace using Principal Component Analysis (PCA) to reduce noise and improve the classifier accuracy. 
In \cite{Damien-2019}, the stream of events is converted to pseudo-frames and features trained on conventional grey level images are transferred to event-based data detection of cars in the observed scene.
In \cite{Chen-2019}, three encoding methods are used to convert the event stream to event frames over a constant time interval. Then, a standard deep neural network with input from the event frames is utilized to predict the locations of pedestrians. 
Aiming to preserve the event temporal structure, in \cite{Cannici-2019}, the conventional convolutional architecture is augmented with an attention mechanism to focus only on relevant events and on the salient spatial portions of frames. 
Object detection by fusing conventional RGB and event streams is explored in \cite{Jiang-2019}. Two kinds of YOLO networks are trained for detecting pedestrians in the RGB and event channels of a DAVIS camera leveraging a confidence map fusion method to improve the accuracy of localization. Construction of the confidence map relies on the construction of pseudo-frames that aggregate events across temporal intervals of predefined length. 
A similar representation is adopted in \cite{Tomy-2022} to address object detection under adverse conditions by using a Feature Pyramid Network to fuse event and RGB data at multiple scales. 
More recently, a cross-modal transformer architecture has been proposed in \cite{Wang-2024} for tracking in joint RGB and event streams. 
In this paper, we propose a transformer-based model, based on the DEtection TRansformer (DETR) architecture \cite{Carion-2020}, which we adapt to study different modality fusion approaches for drone detection.

\custompar{Neuromorphic Drone Detection}
The distinguishing traits of event cameras, characterized by high temporal resolution and high dynamic range, have recently motivated their adoption for detecting drones \cite{Stewart-2021, Stewart-2022, Mandula-2024}. 
In fact, compared to the task of detecting generic objects, detecting drones presents specific challenges due to several unique characteristics. Firstly, drones are often small, making them difficult to detect at a distance or in cluttered environments. Secondly, their high speed allows them to move quickly through a scene, requiring fast and accurate detection methods. Additionally, drones have the ability to hover or remain motionless in the air, which can make them blend into the background and be harder to distinguish from stationary objects. These factors collectively complicate the detection process, necessitating specialized techniques and technologies to reliably identify drones in various scenarios.
In \cite{Stewart-2021, Stewart-2022} a DAVIS camera is used to enable drone detection by seeking the pattern of events generated by the rotating propellers of the drone.
Drone detection is operated by computing a frequency histogram of captured events and using it to feed a classifier capable of distinguishing drone and non-drone airborne objects based on the high-frequency signature and its sub-harmonics.
However, this approach relies on the assumption of a favorable perspective for observing the drone. Observing a drone from a bottom-up perspective allows for clear identification of the propeller blades that are easily distinguishable by an event camera system. However, even with an event camera, capturing the rotation pattern of the propeller blades when the drone is observed from generic points of view is a challenge.
The use of a combined RGB and event camera system for drone detection is investigated in \cite{Mandula-2024} that focuses on the power consumption of an Nvidia Jetson Xavier NX SoM module equipped with a Prophesee EVK4 (events) and a Raspberry Pi (RGB) camera.
The work proposes to train a YOLO-v5 network for the drone detection task. However, details are not provided either about the processing pipeline of the combined RGB and event streams, or about the accuracy of detection.
Similarly, we address the drone detection problem, but we focus on how to effectively leverage multimodal streams through fusion strategies in a transformer-based architecture.


\custompar{Event-based datasets}
Several datasets for event-based object detection have been recently introduced \cite{berlincioni2023neuromorphic, Gehrig-2023, Perot-2020}.
In particular, thanks to the recent widespread availability of high-resolution DVS cameras, many Full HD event-based datasets have been released, such as \cite{innocenti2021temporal, Perot-2020}. Less frequently, event-based datasets are also accompanied by spatio-temporally aligned RGB data, with a few notable examples that leverage multi-modality for more resilient object tracking \cite{Wang-2020}, or to compensate between the RGB and the event domain \cite{Tomy-2022}.
For UAV event-based drone detection, data availability is even more scarce, with drones usually being a relatively small subcategory of bigger datasets as in \cite{wang2024event, wang2023visevent}. 
Finally, only a spoonful of works meets in the intersection of these 3 macro-categories, meaning hybrid DVS-RGB high-resolution datasets with UAV focusing.
Only the dataset recently proposed by Mandula et al. \cite{Mandula-2024}, to the best of our knowledge, have all the desired characteristics but with the caveats of both extremely limited data quantity (only 30 minutes of RGB-event videos) and scenario variety.
In particular, the second point not only may limit the generalizability of methods trained on the dataset, but also hinder the network from adapting to highly complex spatio-temporal scenarios, such as situations in which many moving objects may appear and disappear from the scene at times. 
We propose a novel Neuromorphic-RGB Drone Detection (NeRDD) that comprises 3:30 hours of high-resolution, spatio-temporally aligned Event-RGB recordings of different drones in varying and semantically complex scenarios. Every frame is annotated with drone bounding boxes, enabling tasks such as object detection, tracking and forecasting.

\section{Model Architecture}
\label{sec:model-archit}
We propose a multimodal architecture for drone detection that merges information from neuromorphic and RGB frames.
As a base model, we take inspiration from the DEtection TRansformer (DETR) model~\cite{Carion-2020}, motivated by its flexible modular structure and its effectiveness in common RGB object detection benchmarks.
We first investigate the capabilities of a standard DETR model in both domains individually, highlighting the large performance gap between RGB-based and event-based models.
We then analyze multimodal networks, to bridge the shortcomings of both modalities. To this end, we propose three different event-RGB fusion strategies, ranging from simple pooling layers to more complex attention-based strategies, to retain the best of both worlds and analyze the impact of both modalities.
In the following, we assume to work with event and RGB frames, spatially aligned (i.e., overlappable with little or no misalignment) and temporally synchronized. Event frames $e_i$ are obtained by accumulating all events within temporal intervals of duration $\Delta t=1/F$, where $F$ is the frame rate of the RGB video.
The frame representation of events is made using the camera proprietary API from Propheese\footnote{\url{https://docs.prophesee.ai/stable/guides/frames_generators.html}}.

It must be noted that, in principle, two improvements could be made.
First, the constraint binding the accumulation time $\Delta t$ to the inverse of the frame rate could be relaxed. Doing so would enable a more fine-grained analysis of motion patterns but would increase the computational burden and break the one-to-one pairing between event and RGB frames.
Second, temporal dynamics could be modeled. For simplicity, in this work, we completely disregard time by processing frames individually.
We leave the study of these aspects for future work.

\begin{figure}[t]
     \centering
     \includegraphics[width=\textwidth]{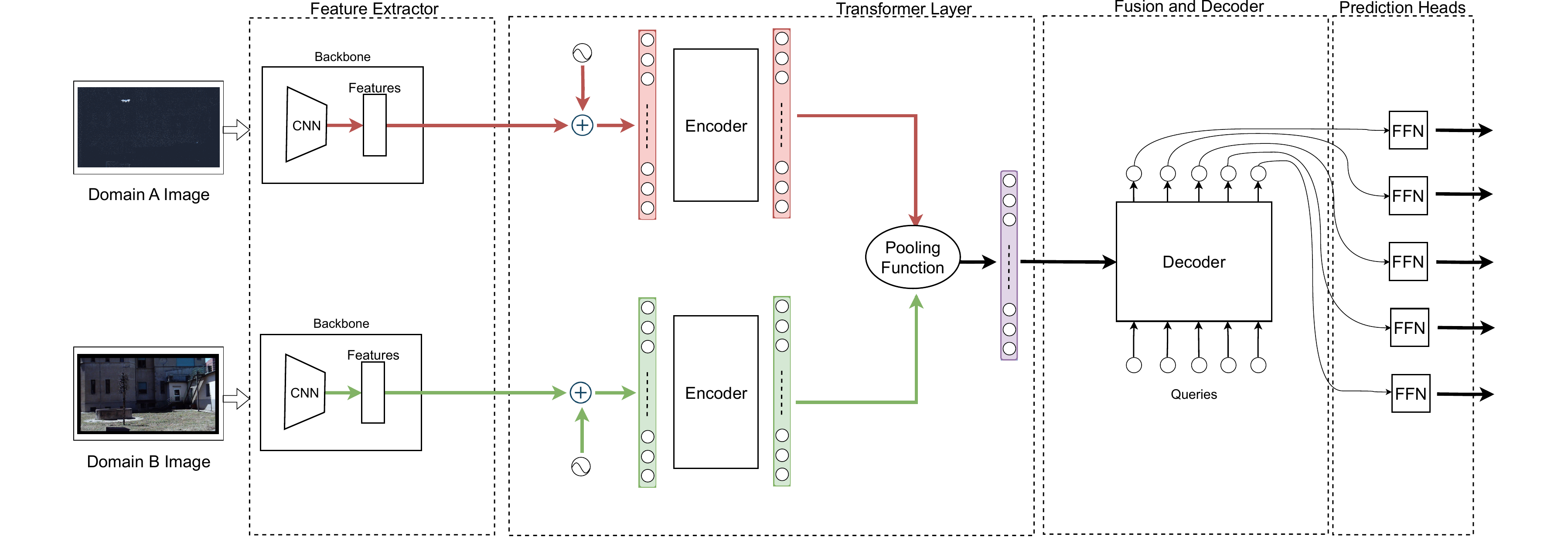}
     \caption{Pooling-based fusion approach. We pool the features after a cut-off layer (the encoder) and process the blended features with the final part of the model.}
   \label{fig:early}
\end{figure}

\custompar{The DETR model} We choose the DETR model as both baseline and foundation to build on for comparing the effectiveness of drone detection by operating on each modality separately and in a multimodal approach, by leveraging on several fusion strategies. DETR combines a CNN backbone for feature extraction (such as Resnet-50 \cite{he2016deep}) and a transformer architecture to detect bounding boxes.
The distinguishing trait of DETR consists of the removal of traditional components like object proposals and anchor boxes with a set of learnable object queries that represent potential objects in the scene. 
The encoder-decoder structure of the transformer allows on the one hand to leverage the image's global context (encoder) and on the other hand to match object queries with spatial tokens (decoder).
The model is trained to perform bipartite matching between queries and objects in the scene through a Hungarian set matching loss. The authors of DETR show that such a loss also removes the need for Non-Maximum Suppression (NMS), as the model learns to filter out overlapping boxes.


\custompar{Fusion Strategies}
We aim to analyze the impact of different data fusion strategies for multimodal drone detection. The solutions we propose mix data at different levels of a common architecture, which is based on DETR.
We propose three groups of fusion strategies: pooling-based fusion, asymmetric modality injection and symmetric fusion.


The \textit{pooling-based fusion} strategy consists in processing the two modalities with a siamese architecture, i.e. replicating the base architecture up to a given cut-off layer. The two partial networks act as feature extractors, that are then blended together via a pooling function and fed to a common head. In principle, any layer of the base model can be used as the cut-off fusion point between modalities and any pooling function capable of preserving the shape of the features can be used. In our experiments, we use the channel-wise average pooling function.
Thanks to the preserved spatial structure in convolutional features and in transformer tokens, regions with low activation values are discarded, preserving instead the regions in which both modalities agree. 
As a downside, if convolutional maps are pooled together, the local receptive field of convolutions requires the data in the two modalities to be spatio-temporally aligned for the model to be effective.
In Fig. \ref{fig:early} we show our multimodal architecture with the pooling performed after the encoder layers.

\begin{figure}[t]
     \centering
     \includegraphics[trim={50 0 50 0},clip,width=\textwidth]{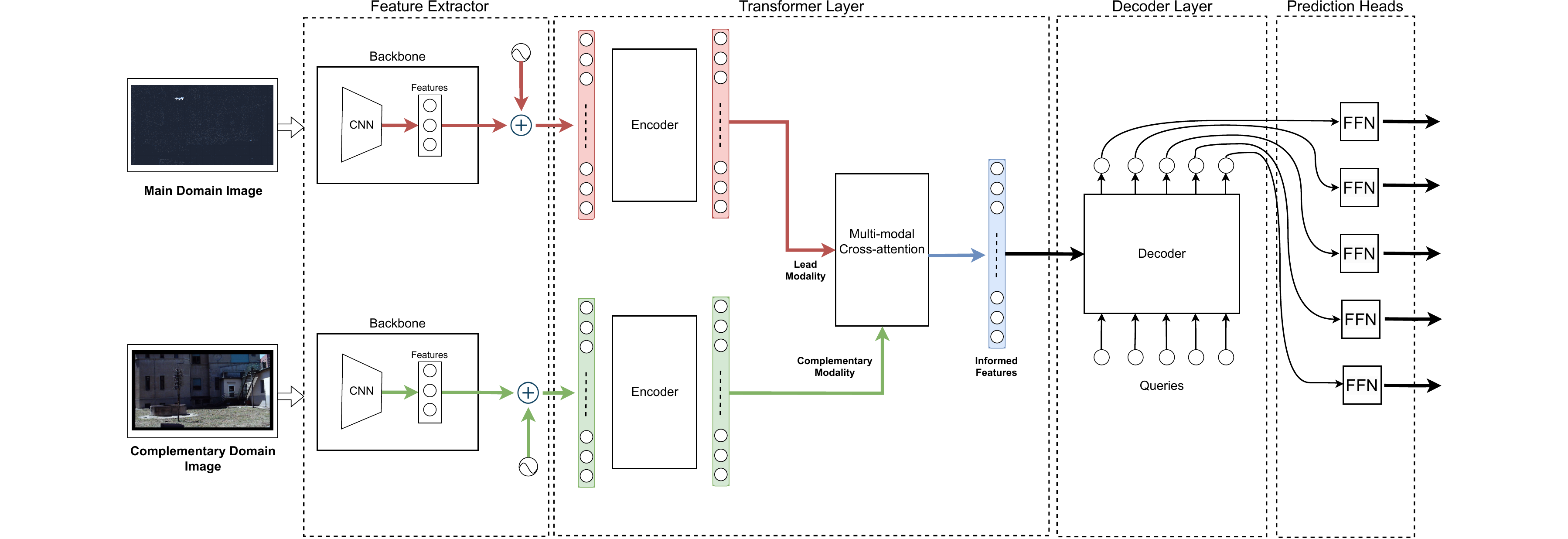}
     \caption{Asymmetric modality injection. The main domain (event) is informed about the complementary domain (RGB) thanks to a cross-attention mechanism that blends the features asymmetrically.}
     \label{fig:middle}
\end{figure}

The \textit{asymmetric modality injection}, relies on a one-way information injection, meaning that only one modality at a time will be informed about the other. In particular, the architecture is enriched with a cross-modality cross-attention layer inside the transformer module.
In the cross-attention mechanism, the main modality is projected into a query matrix, whereas the complementary modality generates keys and values. The rationale behind this architectural choice is that tokens belonging to the main modality are used to extract meaningful appearance information by attending the complementary domain.
The tokens obtained via such multimodal cross-attention are then added to the original tokens of the main domain via a skip-connection to preserve the original information.
In Fig.~\ref{fig:middle} we depict the architecture with asymmetric modality injection using event data as the main domain and RGB as the complementary domain. 

\begin{figure}[t]
     \centering
     \includegraphics[width=.95\textwidth]{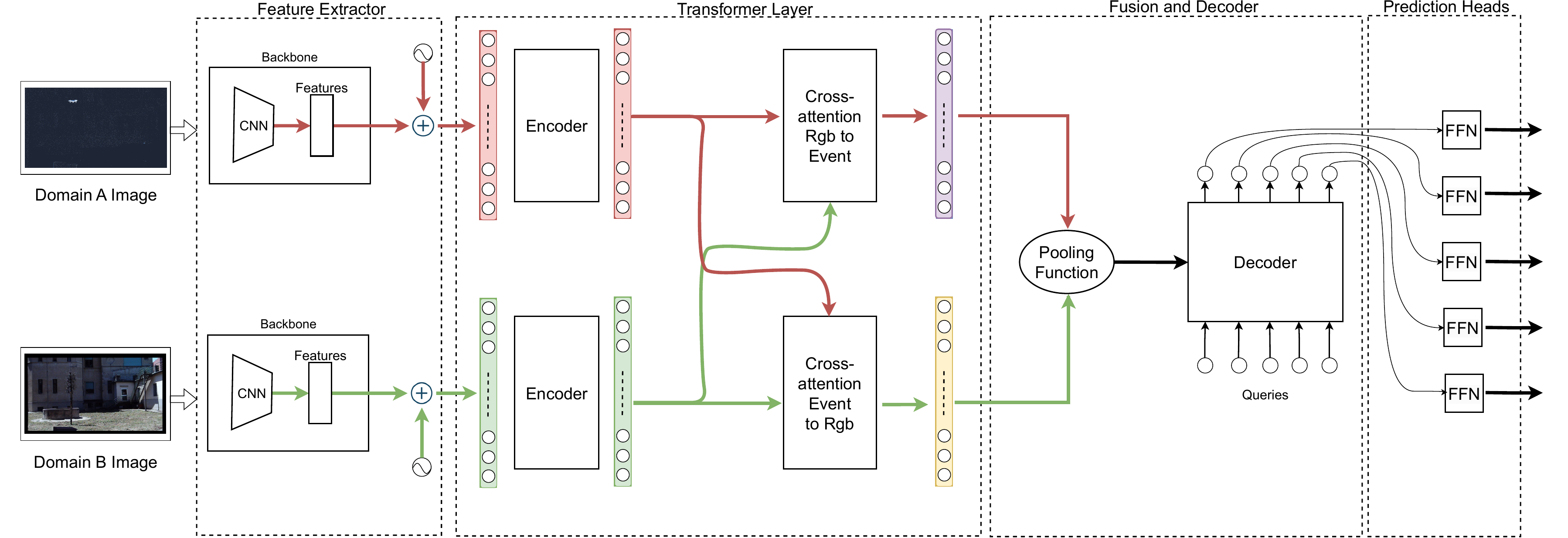}
     \caption{Symmetric fusion architecture. Two asymmetric injections are performed, to inform the two modalities of each other. A final pooling layer is used to merge the features symmetrically.}
     \label{fig:doublecross}
\end{figure}

Finally, \textit{symmetric fusion} builds on top of the previous modalities by applying two parallel asymmetric modality injections (swapping the main and the complementary domains).
In this way, both modalities are separately informed about the other, after which they are once again pooled together to return a single set of features to be passed on to the final transformer decoder.
Fig.~\ref{fig:doublecross} shows our symmetric fusion architecture.

\section{Neuromorphic-RGB Drone Detection Dataset}
\label{sec:dataset}

In the following we provide an overview of the NeRDD dataset, our novel dataset for neuromorphic-RGB drone detection.

\custompar{Technical details}
To collect the data, a double-camera single-mount setup has been made, placing side-by-side an HD DVS camera and a standard RGB camera. 
In particular, the event-based camera is a Prophesee EVK4 HD, equipped with an 8mm optic and a Sony IMX636ES sensor. The RGB camera is a Svpro HD camera with varifocal lens and a resolution up to 1280x720 at 30fps.
Since the distance between the cameras is relatively small with respect to the target depth in the scene, we can safely assume that the resulting captured images will be overlappable given a minor shift in the x-axis. 
This gives the chance to project coordinates between the RGB-Event domains. 

The recorded drones are 2 quadcopters of different dimensions and flying dynamics. 
The first is a DJI Tello EDU quadcopter; it weighs 87g and has dimensions of 98×92.5×41 mm. These physical characteristics make the drone highly susceptible to atmospheric phenomena, such as sudden variations in wind direction and speed. The limited dimensions also make the detection task particularly challenging, depending on the drone's distance from the camera. 
The second drone is a DJI Mini 2, another quadcopter of more prominent dimensions. While structurally similar to the Tello EDU drone, its dimensions and weight (245×290×55 mm for 250g) make the drone much more stable even in complicated atmospherical conditions, while also hindering its ability to swift and perform sudden movements. 

\begin{table}[t]
\begin{center}
\caption{Comparison of existing event-based drone datasets. Other datasets either have a low resolution, do not contain RGB versions of the samples, are not Drone-centric or are very small.}
\label{table:dataset_comparison}
\begin{tabular}{l|cccc}
\textbf{Dataset}~ & ~\textbf{Resolution}~ & ~\textbf{RGB/Event}~ & ~\textbf{Hours}~ & ~\textbf{Drone-Centric}\\
\hline
\textbf{VisEvent\cite{wang2023visevent}} & 346 x 260 & \checkmark/\checkmark & <5 & $\times$\\
\hline
\textbf{EventVOT\cite{wang2024event}}~ &  ~1280 x 720 (HD) & $\times$/\checkmark & <5 & $\times$\\
\hline
\textbf{F-UAV-D\cite{mandula2024towards}} & ~1280 x 720 (HD)& \checkmark/\checkmark & 0.5 & \checkmark\\
\hline
\textbf{NeRDD}(Ours) & ~1280 x 720 (HD) & \checkmark/\checkmark & 3.5 & \checkmark \\
\end{tabular}

\end{center}
\end{table}

\custompar{Dataset preprocessing}
To enable a pixel-wise overlapping between event and RGB data, and thus obtain projectable coordinates between the two domains, we adopted a precise pipeline for spatio-temporal synchronization.
The dataset comes with pre-generated event frames at 30 fps, to match the RGB camera recording speed. Using these frames we achieved temporal synchronization aligning the RGB and event frames to match a displayed chronometer, recorded at the beginning of each video\footnote{The initial part has been manually removed in post-processing.}. Once we achieved temporal synchronization, we also calibrated the cameras, obtaining intrinsic parameters, which we used to undistort the frames. We also crop and pad the RGB frames to spatially match the event domain.
To get the ground truth bounding boxes for drone locations, we used a semi-automatic annotation pipeline.
We relied on the ready-for-use metavision spatter tracking script\footnote{\url{https://docs.prophesee.ai/stable/samples/modules/analytics/tracking_spatter_py.html}} to identify moving event blobs to obtain an initial set of boxes. Then, a round of manual inspection was made to ensure the ground truth correctness and amend annotation errors, add missing boxes and remove the large amount of blobs not representing drones. Finally, a simple interpolation between temporally adjacent boxes was performed to ensure smooth and continuous annotations.

The resulting dataset comprises 3.5 hours of multimodal recordings (a total of 7 hours of footage) at 30 FPS, divided into 115 different videos. Location and background activity significantly varies across videos. Both modalities have HD resolution ($1280 \times 720$). A comparison of the NeRDD dataset with other event-based drone datasets is given in Tab. \ref{table:dataset_comparison}. A few existing datasets contain a small amount of drone footage along with other categories. The only existing neuromorphic drone-centered dataset is F-UAV-D\cite{mandula2024towards}, which has a temporal extent 7 times smaller than ours and exhibits reduced variability in the recordings.
We summarize the characteristics of NeRDD in Tab. \ref{table:dataset_stats}. We chose to also keep in the dataset a subset of frames where drones are not visible. This is due to the fact that a drone can exit the field of view and re-enter a few frames later. The presence of empty frames poses a challenge for a detector as false positives might be detected and also for detection/tracking methods that leverage temporal information and that may be explored in future works.
The dataset is publicly available at the following URL\footnote{URL to be inserted in the camera ready to preserve anonymity}. We show in Fig. \ref{fig:dataset_samples} spatiotemporally aligned event and RGB samples with drone bounding boxes.

\newcommand{\figwidth}{.43\linewidth}
\begin{figure}[t]
\centering
    \includegraphics[trim={50 50 50 50},clip,width=\figwidth]{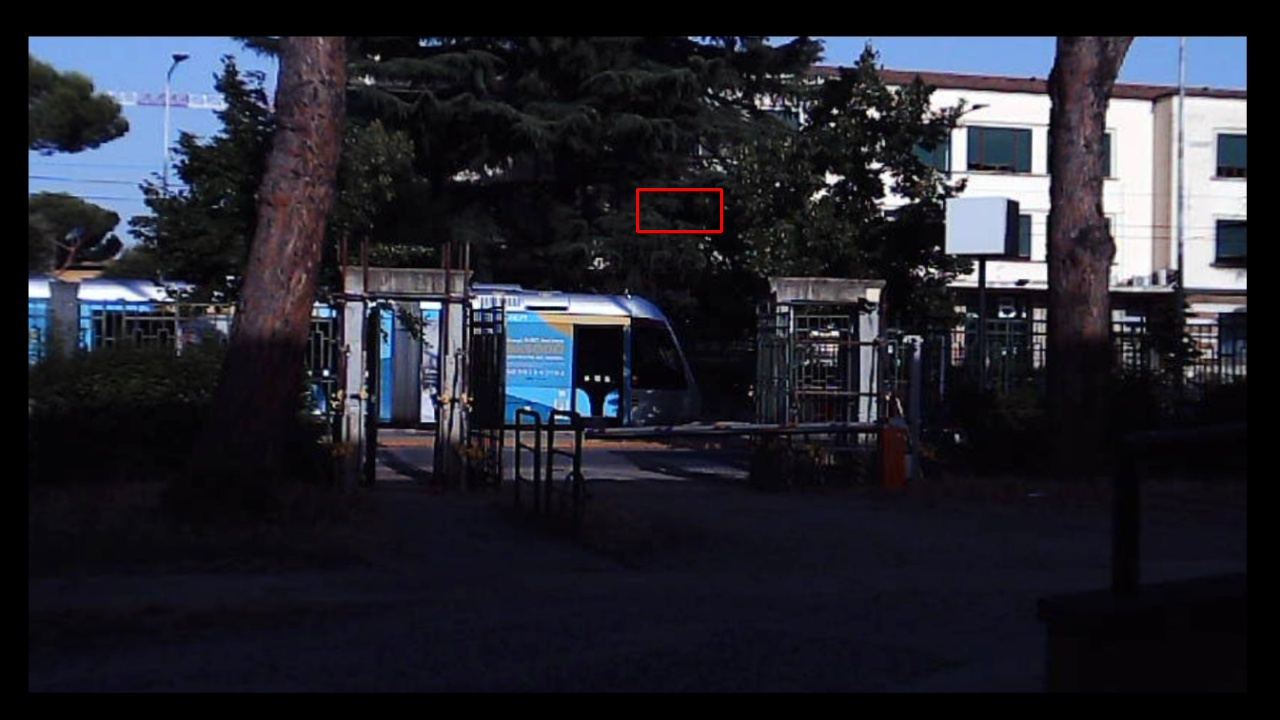}
    \includegraphics[trim={50 50 50 50},clip,width=\figwidth]{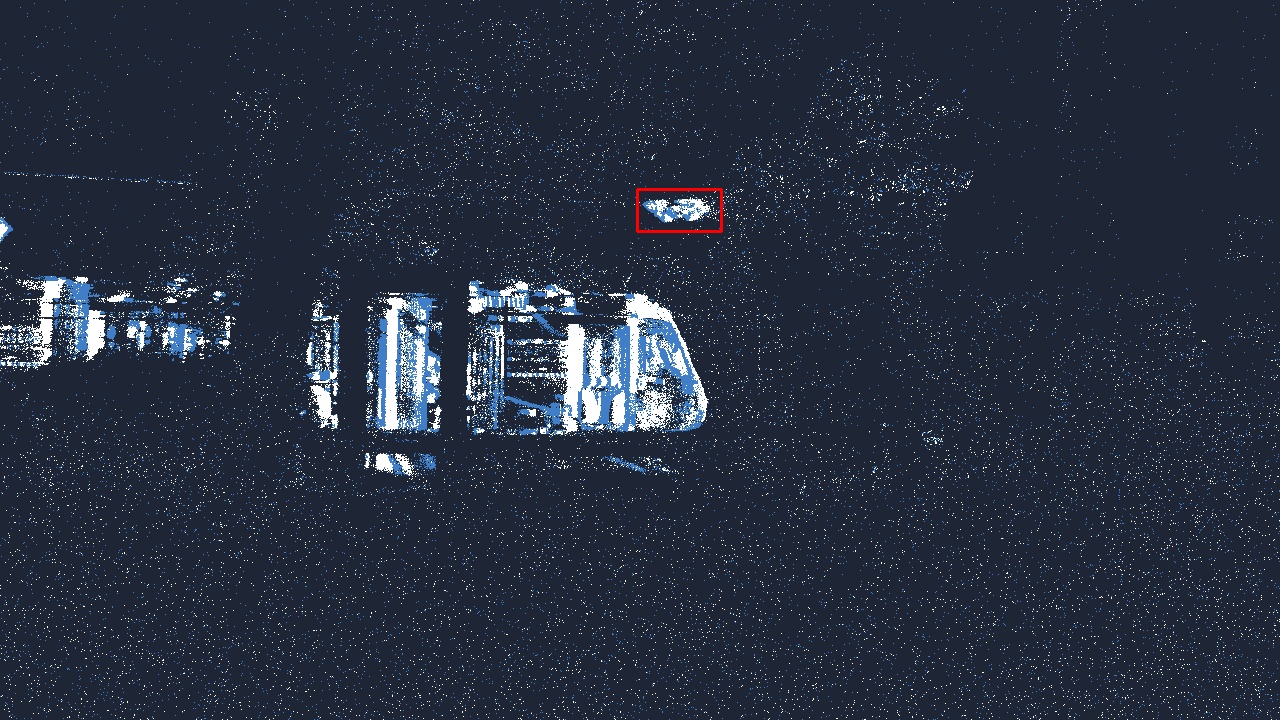} 
    \includegraphics[trim={50 50 50 50},clip,width=\figwidth]{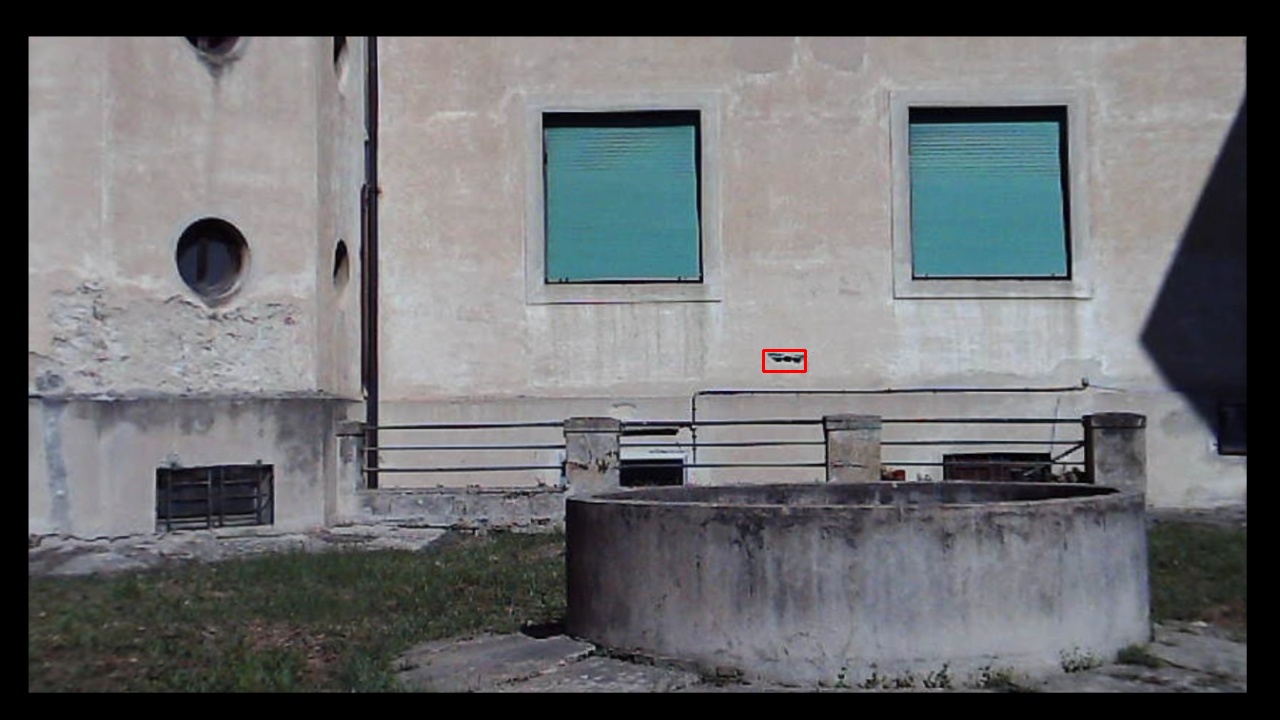}
    \includegraphics[trim={50 50 50 50},clip,width=\figwidth]{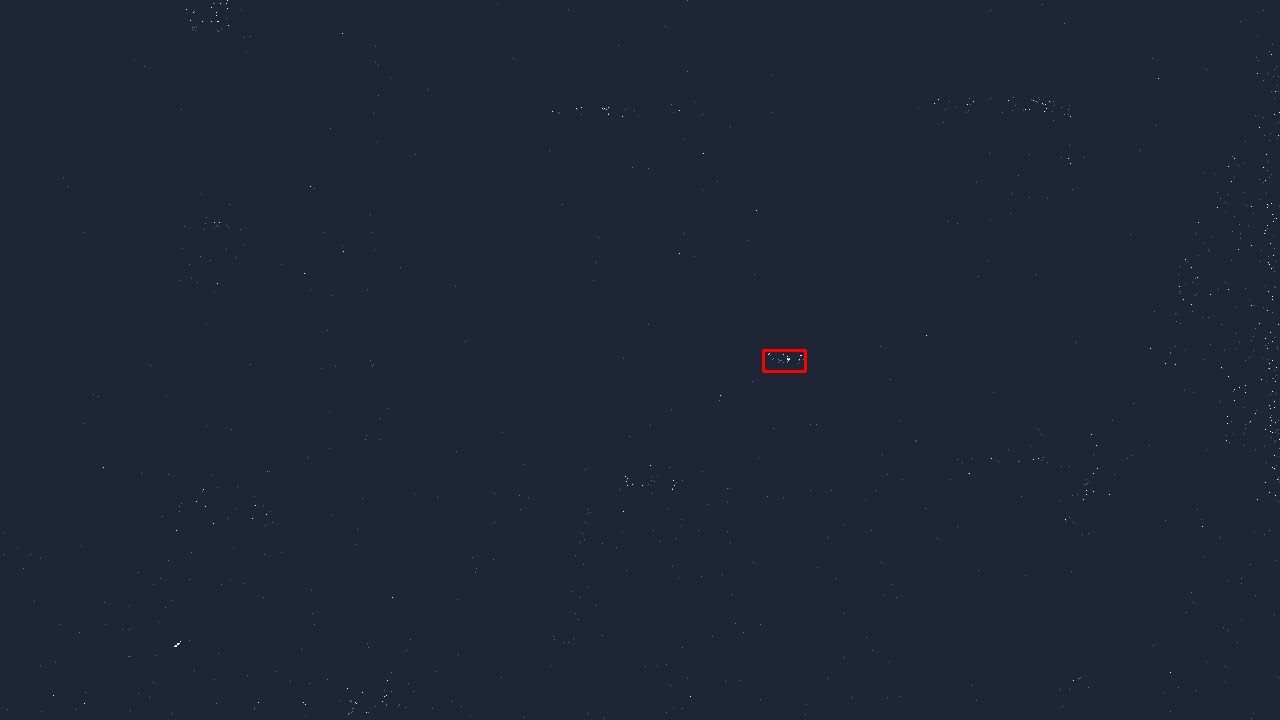} 
    \includegraphics[trim={50 50 50 50},clip,width=\figwidth]{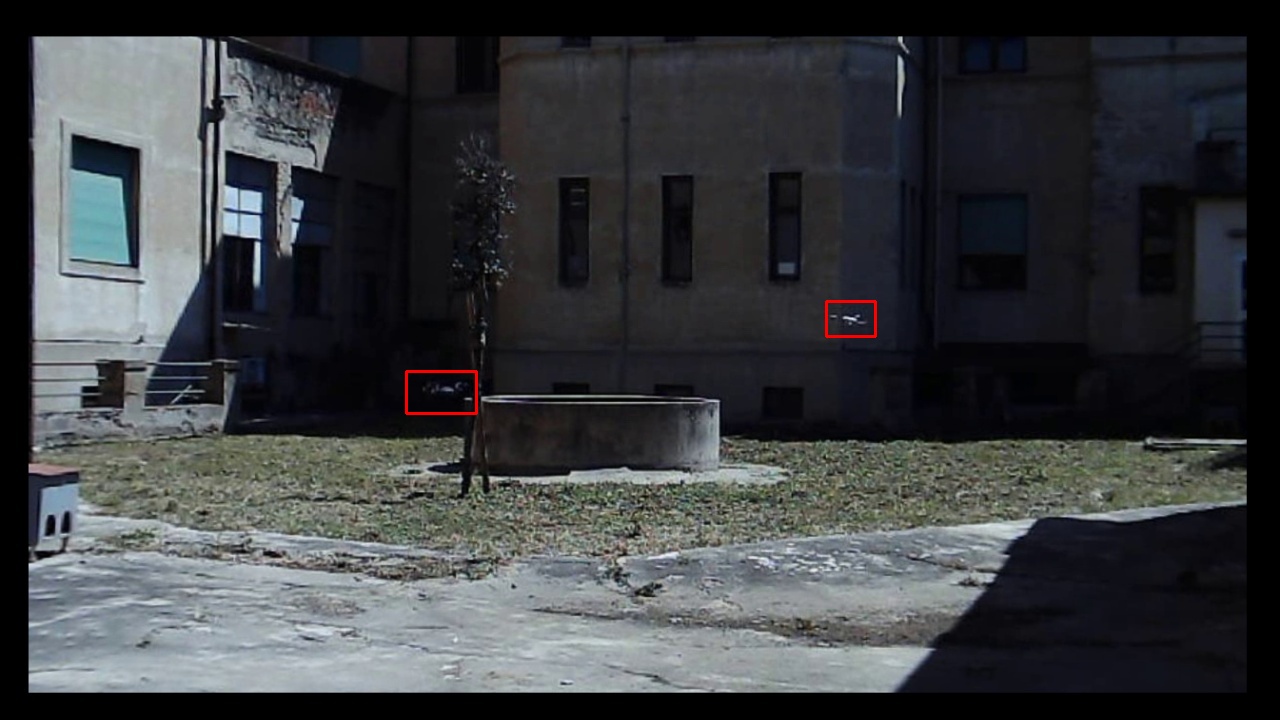}
    \includegraphics[trim={50 50 50 50},clip,width=\figwidth]{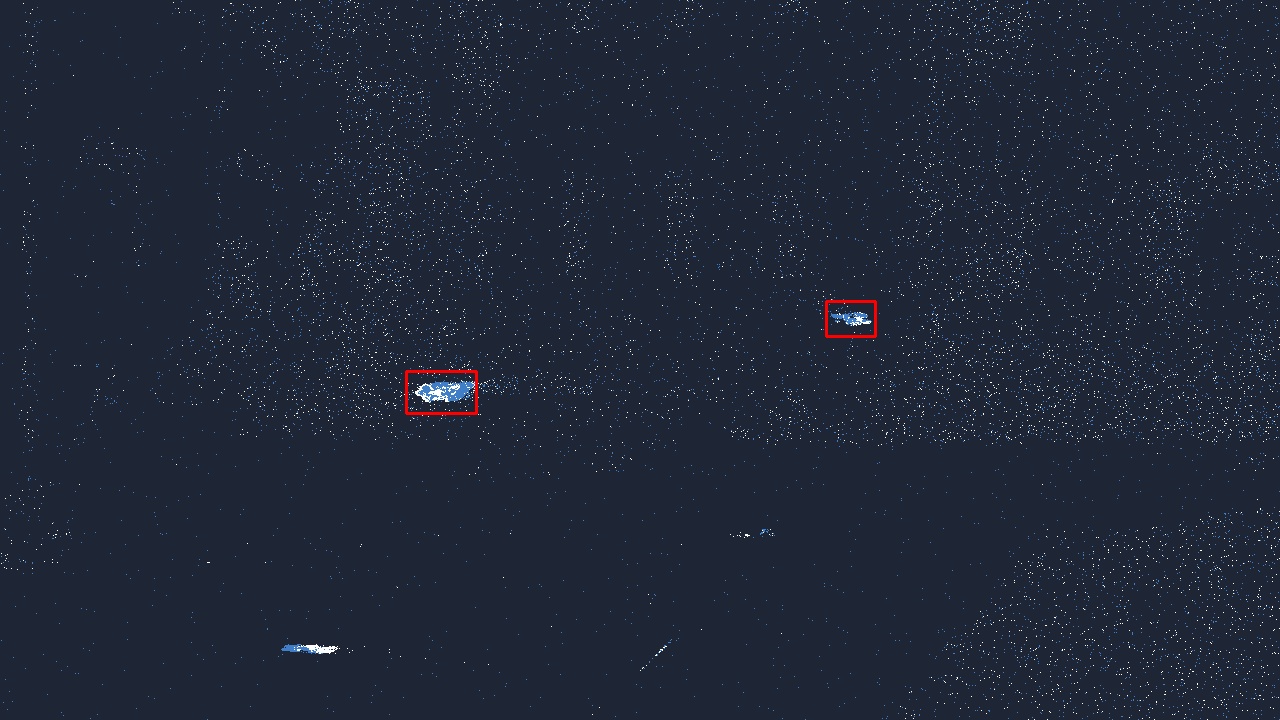} 
    \includegraphics[trim={50 50 50 50},clip,width=\figwidth]{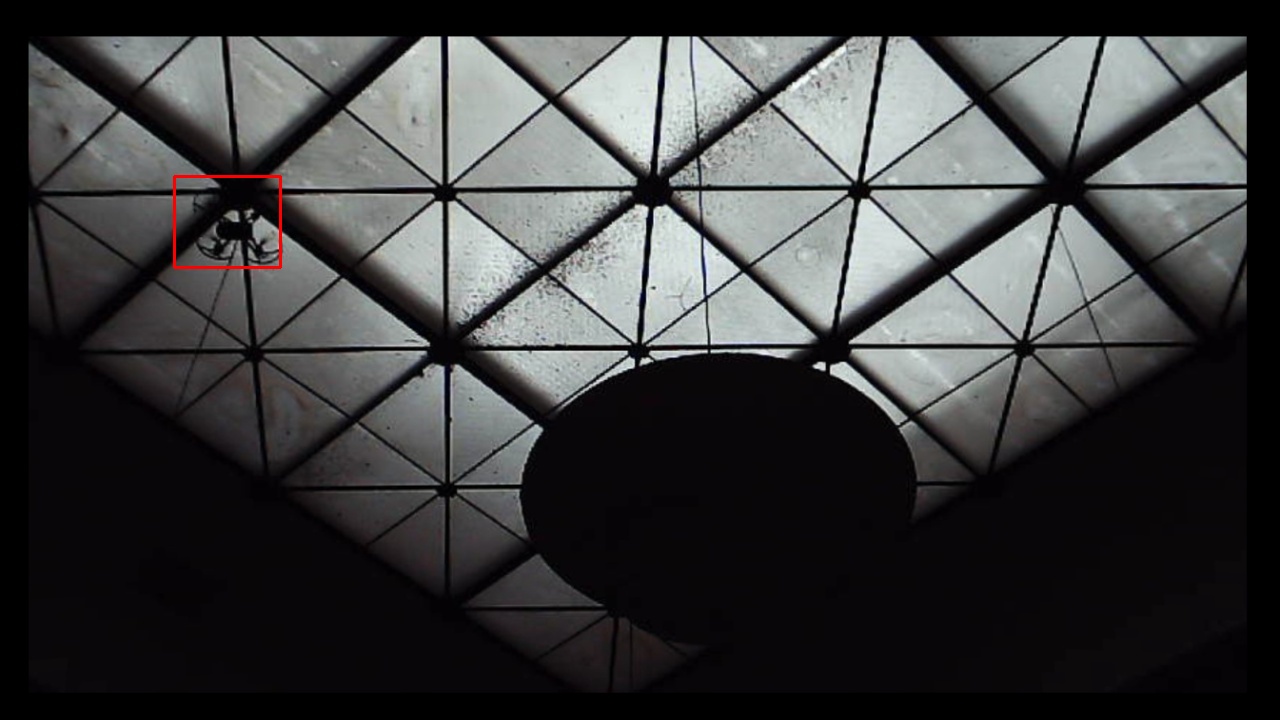}
    \includegraphics[trim={50 50 50 50},clip,width=\figwidth]{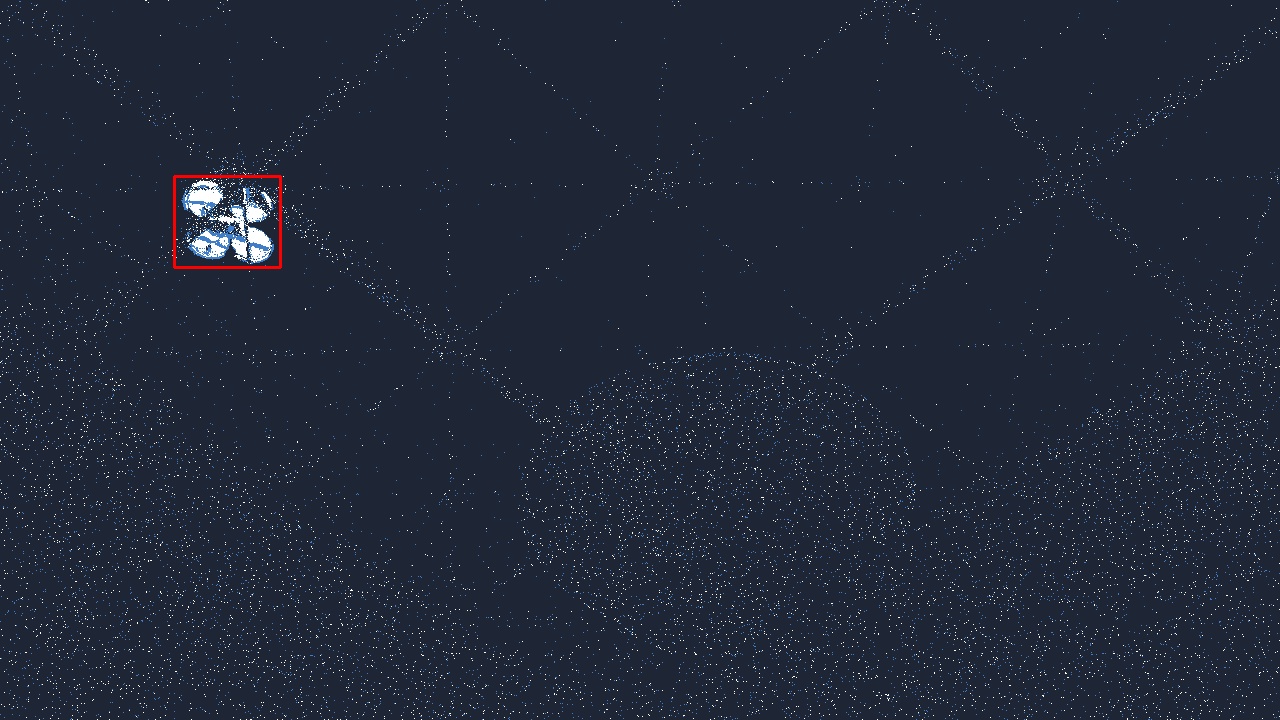} 
\caption{Example of aligned frames in the dataset. The left figures are the RGB images, while the right images are the event frames obtained by accumulating all the events in a time slice of 33ms. The red boxes in the images are the ground truth bounding boxes for the detection, with the same identical coordinates for both domains. Crops of the original images are displayed for better visualization.}
\label{fig:dataset_samples}
\end{figure}

\begin{table}[t]
\begin{center}
\caption{Principal characteristics of the NeRDD dataset. The number of frames refers to a single modality, i.e. the complete number of frames doubles when taking into account both modalities.}
\label{table:dataset_stats}
\resizebox{\textwidth}{!}{
\begin{tabular}{c|c|c|c|c|c|c}
 \textbf{Frames} & \textbf{Frames w/ Drone} & \textbf{Frames w/o Drone} & \textbf{Tot length} & \textbf{Videos} & \textbf{FPS} & \textbf{Resolution} \\
\hline
2 $\times$ 382.545 & 294.490 ($\approx$80\%) & 88.055 ($\approx$20\%) & 3h:35m & 115 & 30 & 1280x720\\
\end{tabular}
}

\end{center}
\end{table}

\section{Experimental Results}
\label{sec:experimental}
In this section, we report the results of our drone detection architectures, detailing also the experimental setting. 

 \subsection{Implementation Details}
\label{sec:implementationdetails}
We divided the dataset into train and test data, following a video-wise 80/20 split (92 videos for train, 23 for test) to avoid similar frames in both splits.
All models have been fine-tuned on a pre-trained DETR model, changing the number of object queries and the classification head. In particular, since only a maximum of 2 contemporary drones are present in each frame, we opted to lower the number of object queries from 100 to 5. 
All the models have been trained for 30 epochs with a learning rate of $1e^{-5}$ with a decay of an order of magnitude every 15 epochs. The optimizer is AdamW as in standard DETR and the batch size is 8.

\subsection{Evaluation}
\label{sec:evaluation}

\newcommand{\figreswidth}{.2\linewidth}
\begin{figure}[t]
\begin{center}
\begin{tabular}{ccccc}
\textbf{DETR (RGB)}&\textbf{DETR (EV)}&\textbf{EV-to-RGB}&\textbf{RGB-to-EV}&\textbf{Pooling (Enc)}\\
\includegraphics[trim={250 50 50 50},clip,width=\figreswidth]{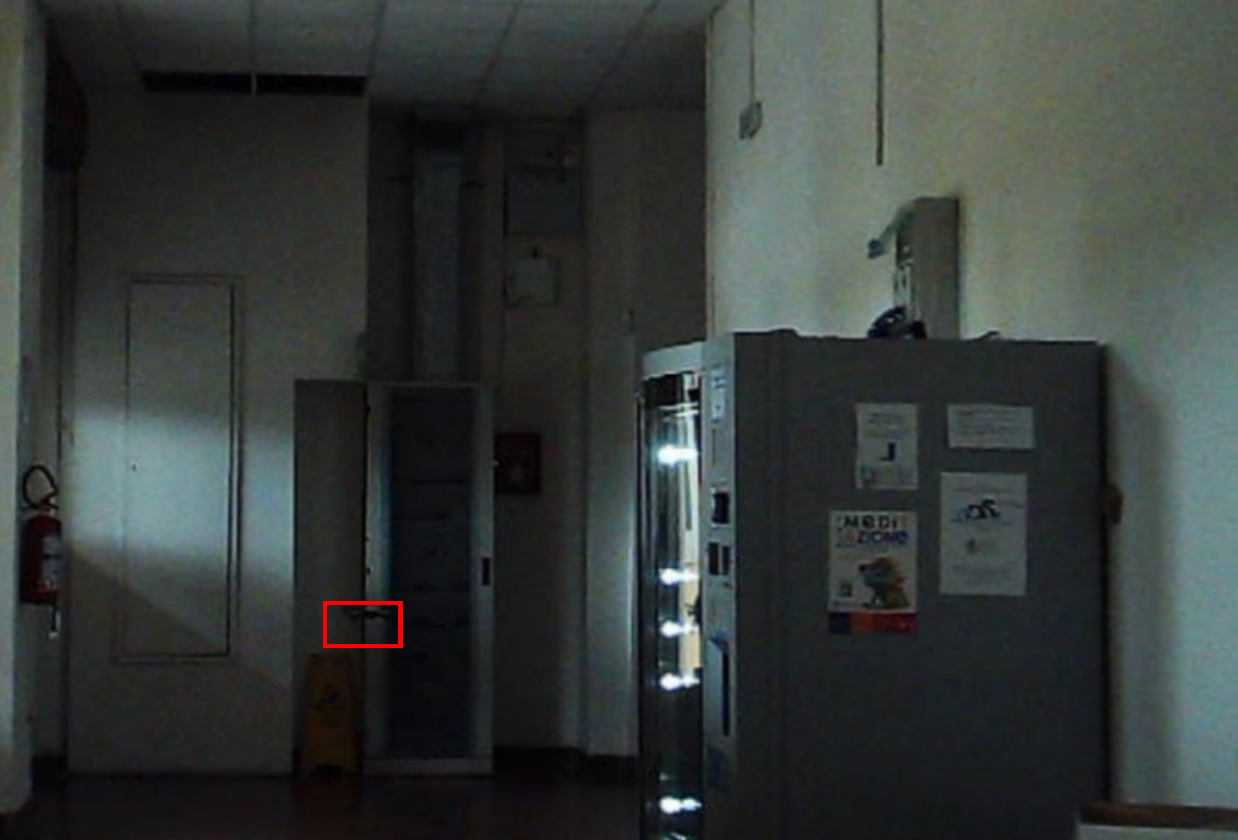} & \includegraphics[trim={250 50 50 50},clip,width=\figreswidth]{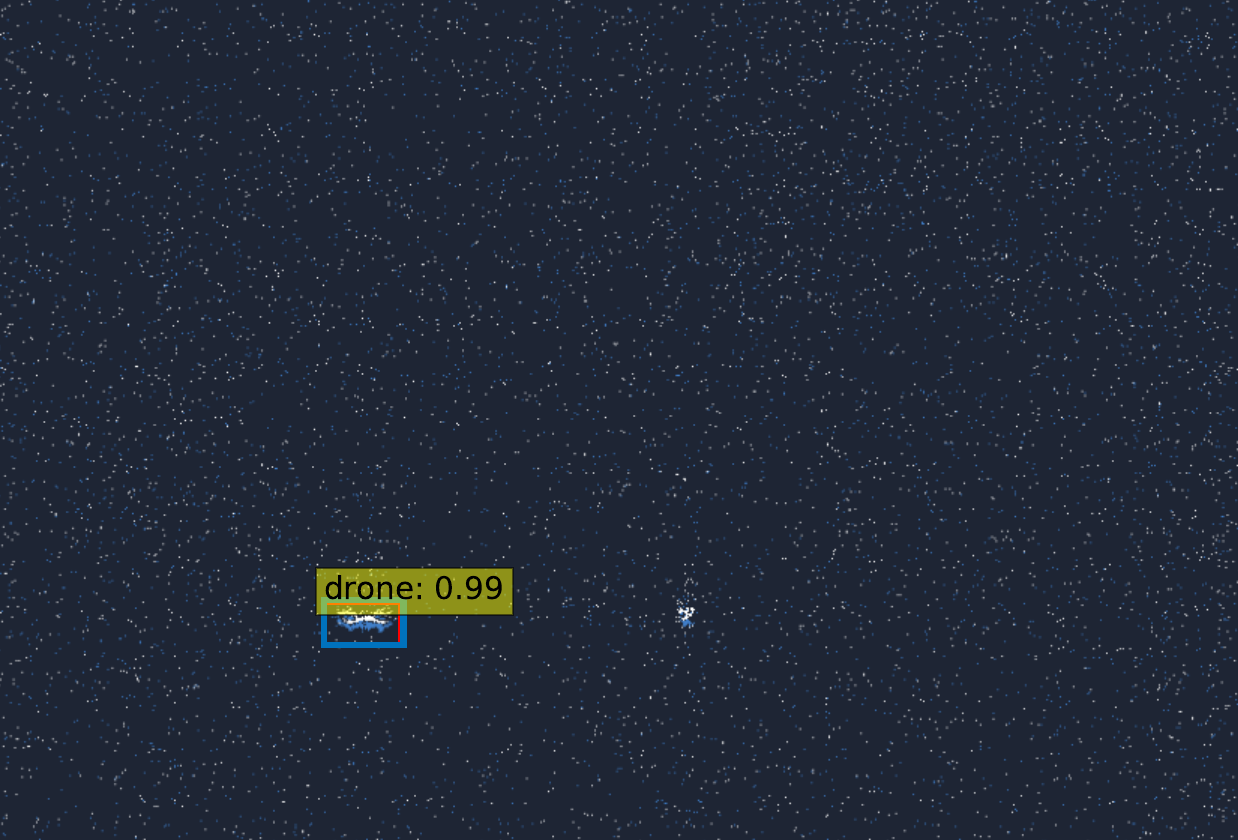}  & \includegraphics[trim={250 50 50 50},clip,width=\figreswidth]{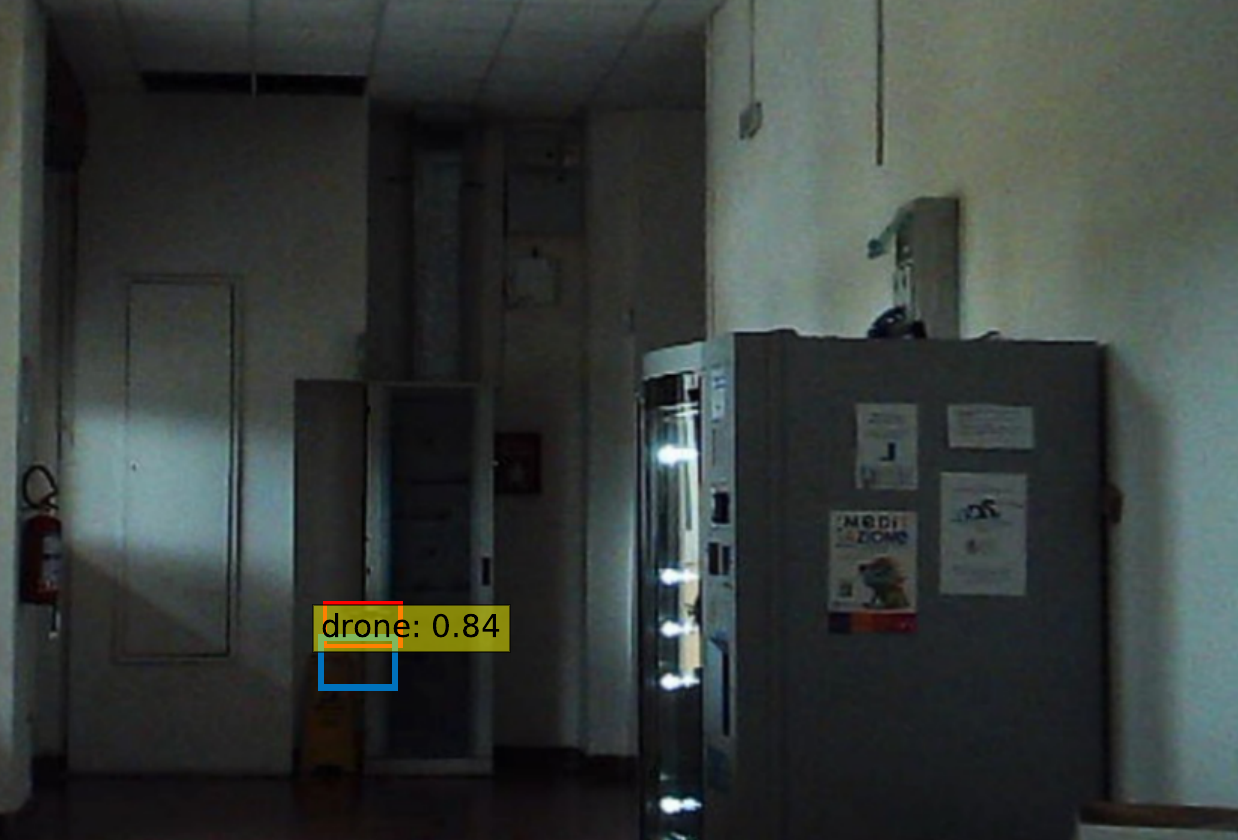} & \includegraphics[trim={250 50 50 50},clip,width=\figreswidth]{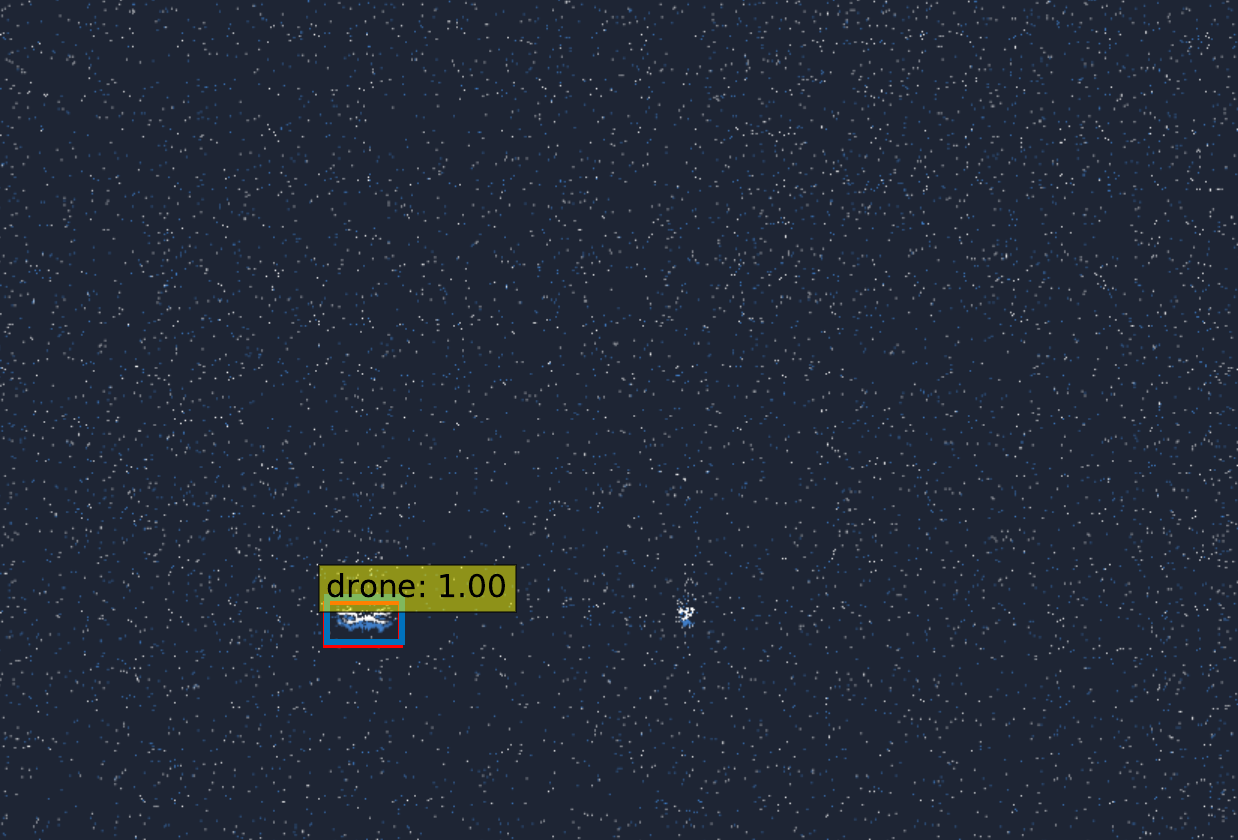} & \includegraphics[trim={250 50 50 50},clip,width=\figreswidth]{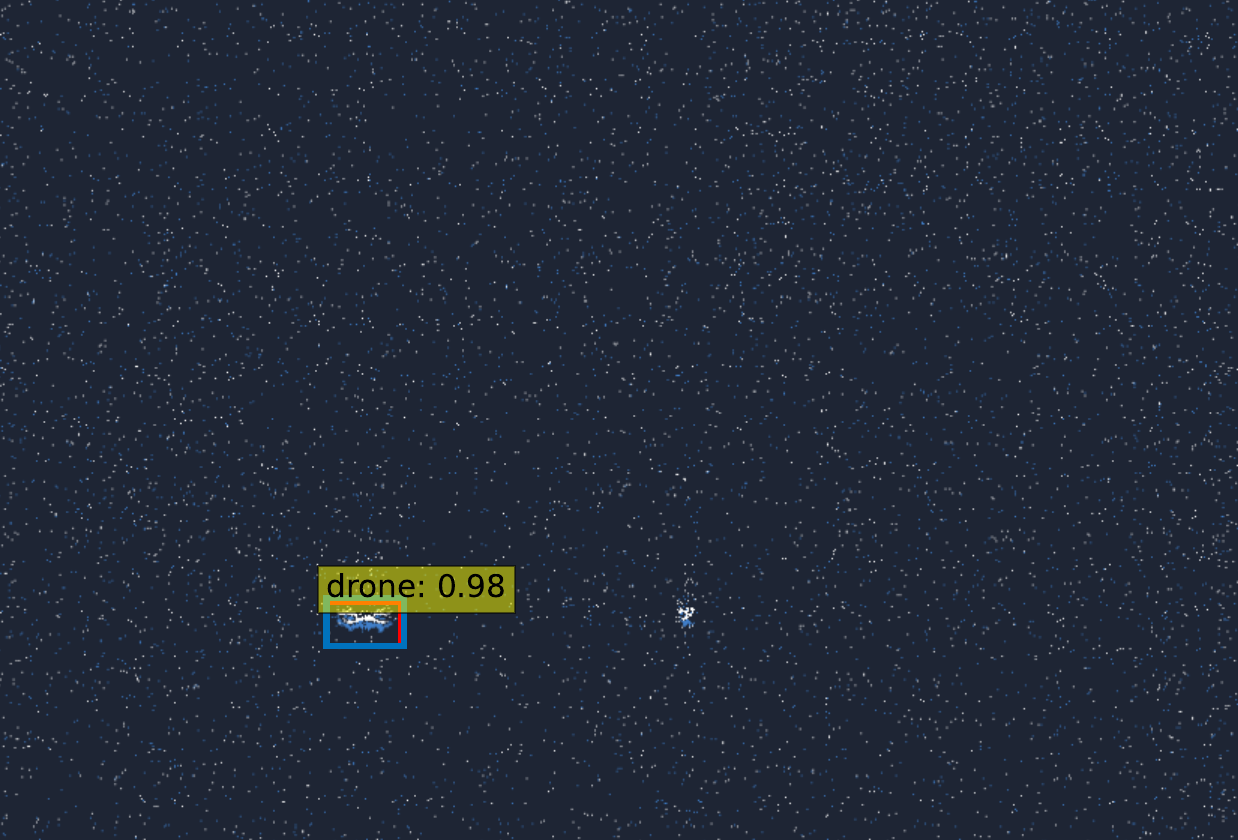}\\
\includegraphics[trim={250 50 50 50},clip,width=\figreswidth]{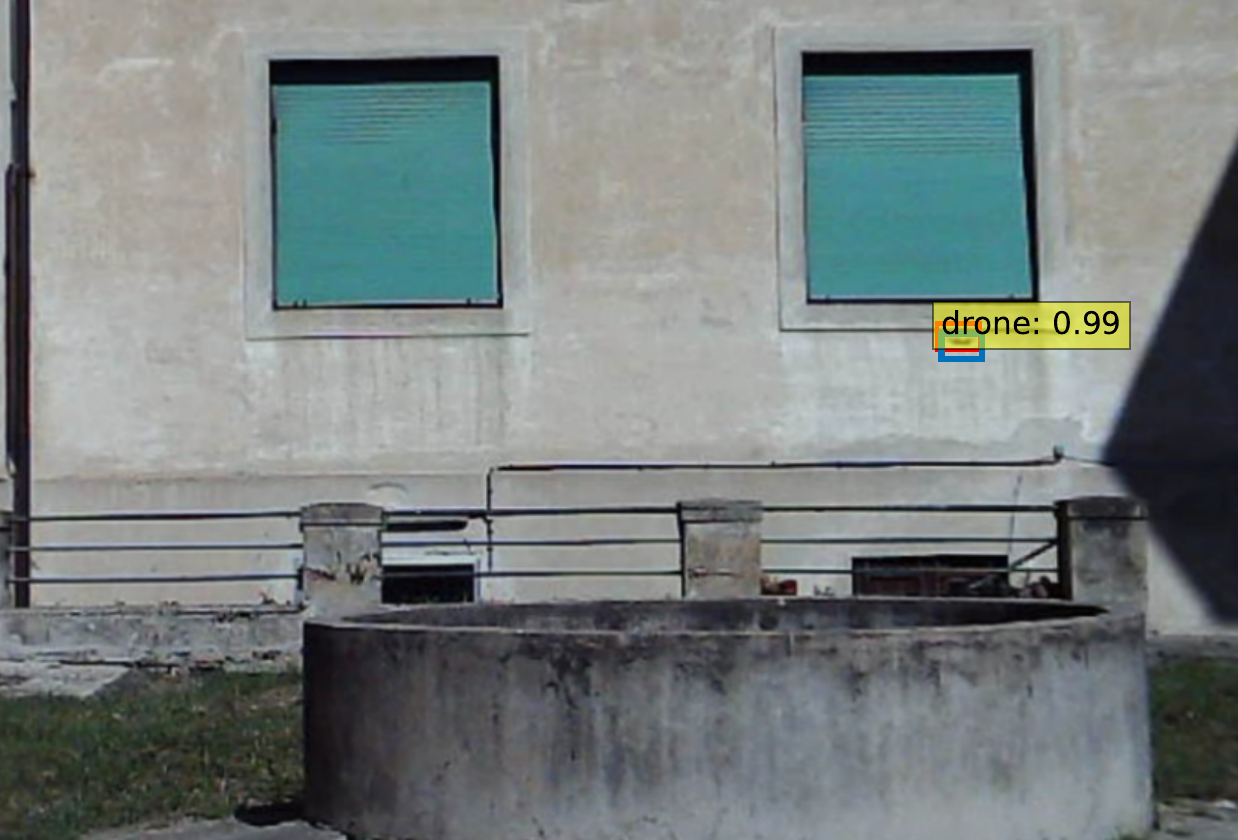} & \includegraphics[trim={250 50 50 50},clip,width=\figreswidth]{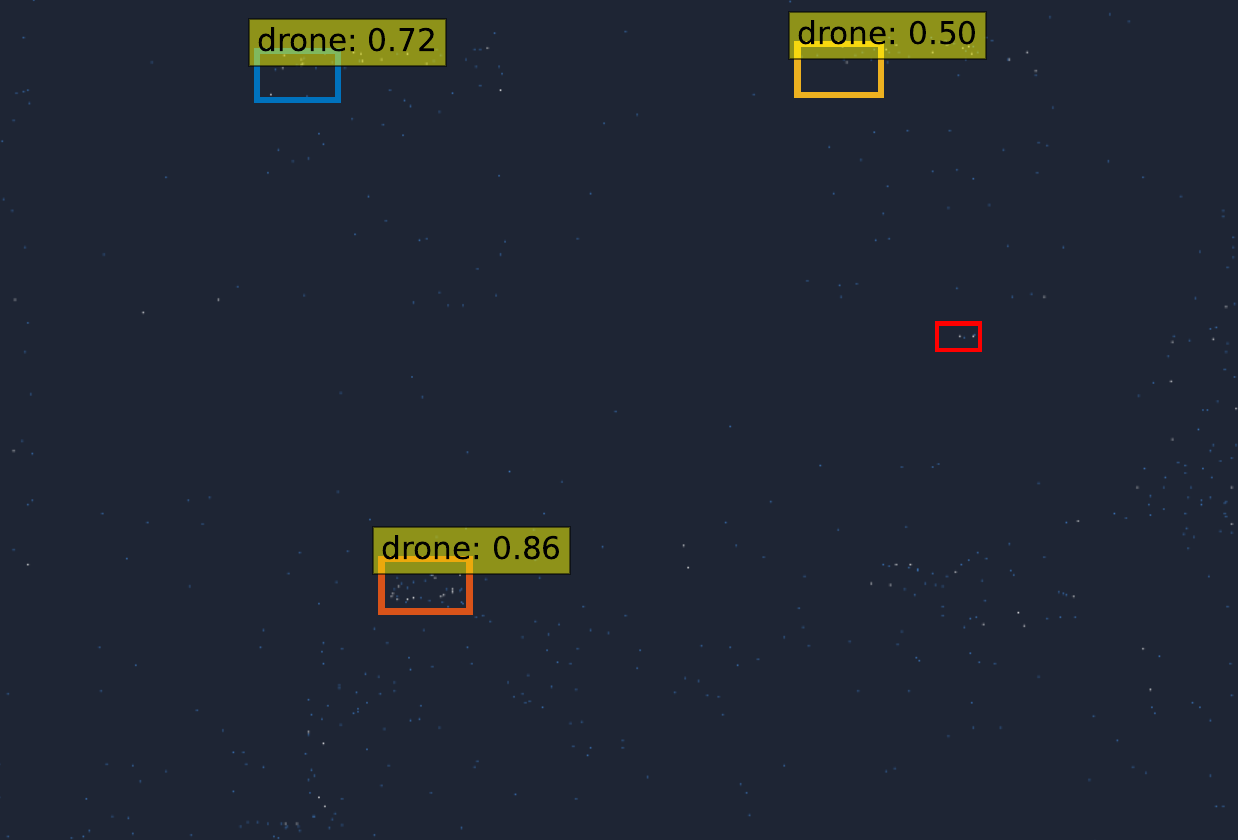}  & \includegraphics[trim={250 50 50 50},clip,width=\figreswidth]{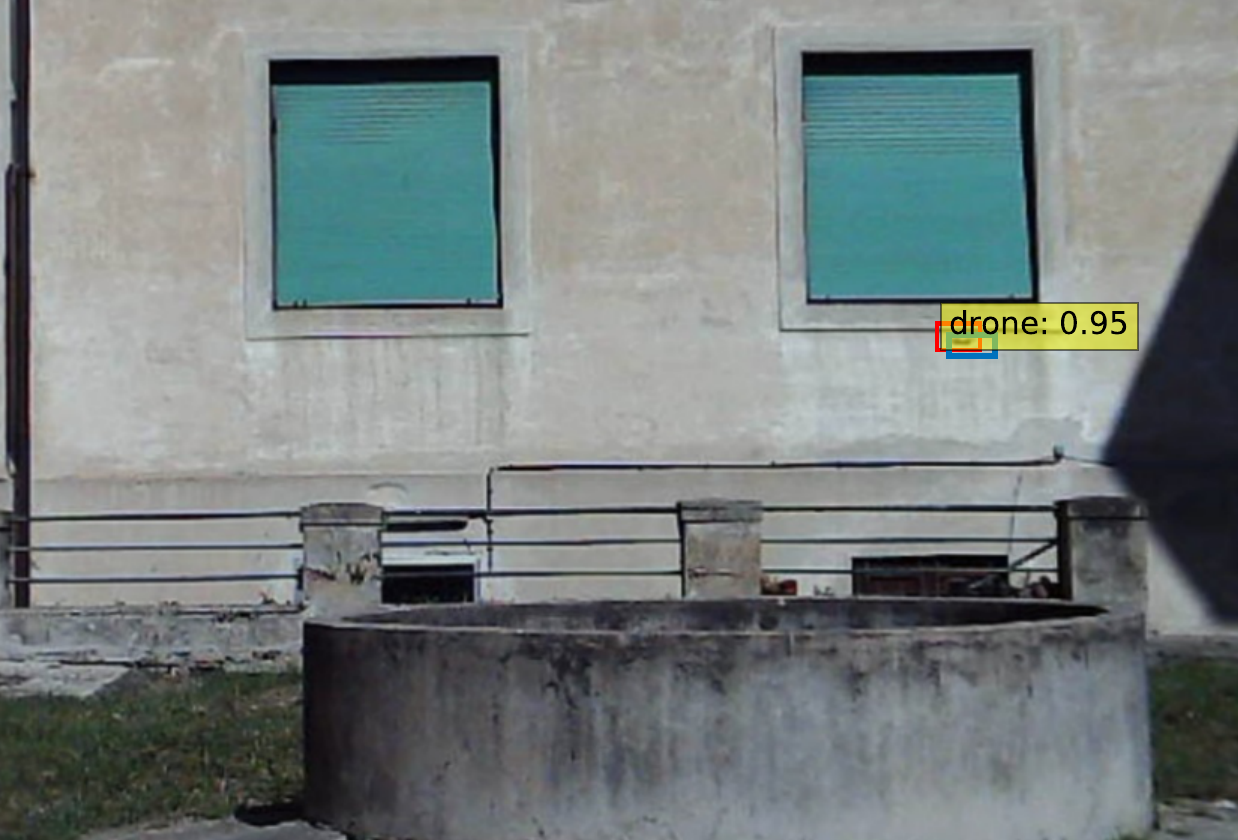} & \includegraphics[trim={250 50 50 50},clip,width=\figreswidth]{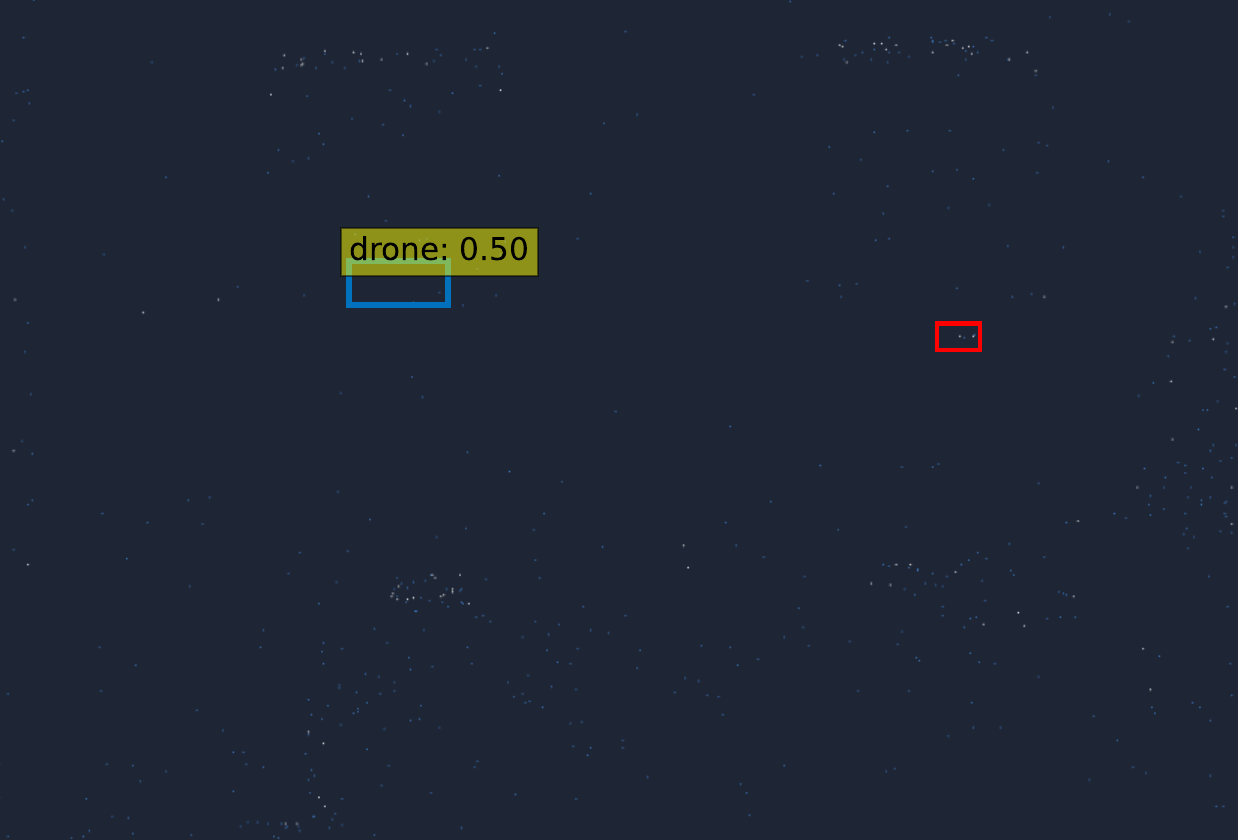} & \includegraphics[trim={250 50 50 50},clip,width=\figreswidth]{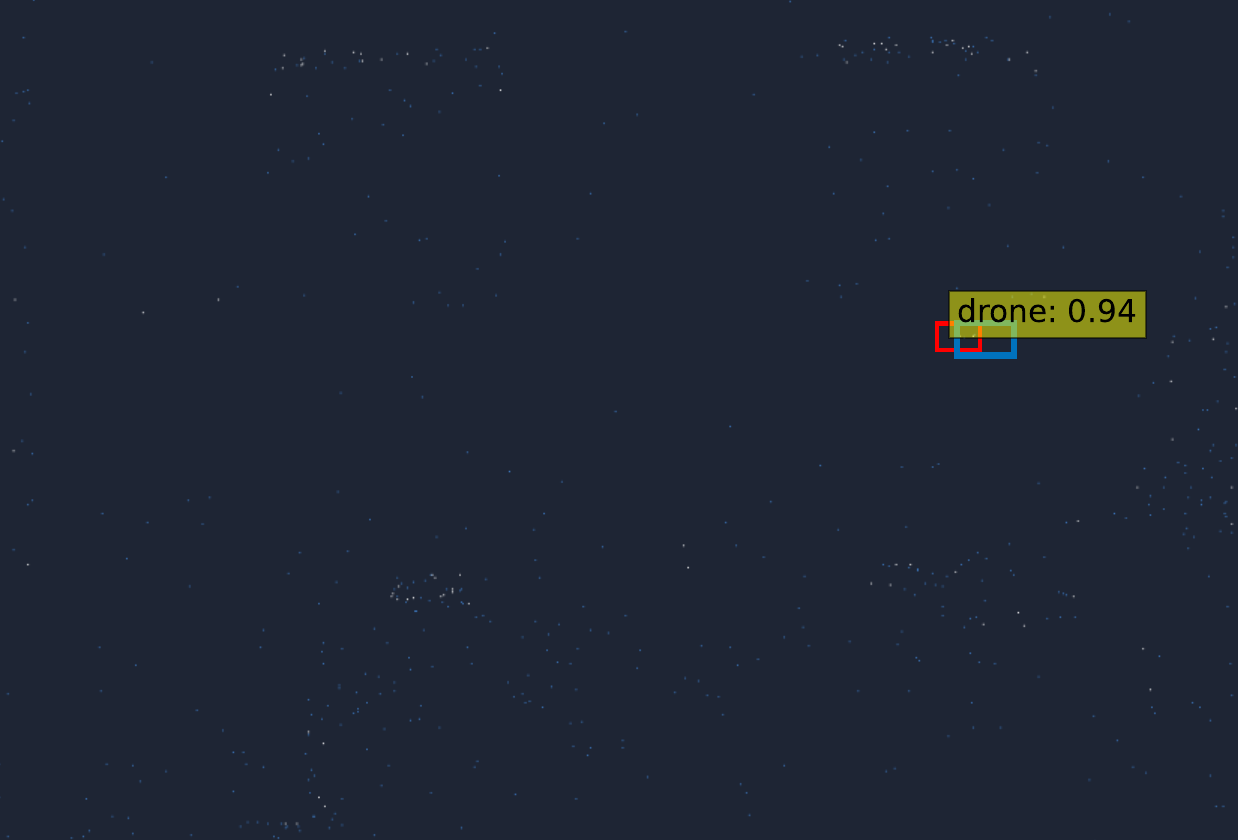}\\
\includegraphics[trim={250 50 50 50},clip,width=\figreswidth]{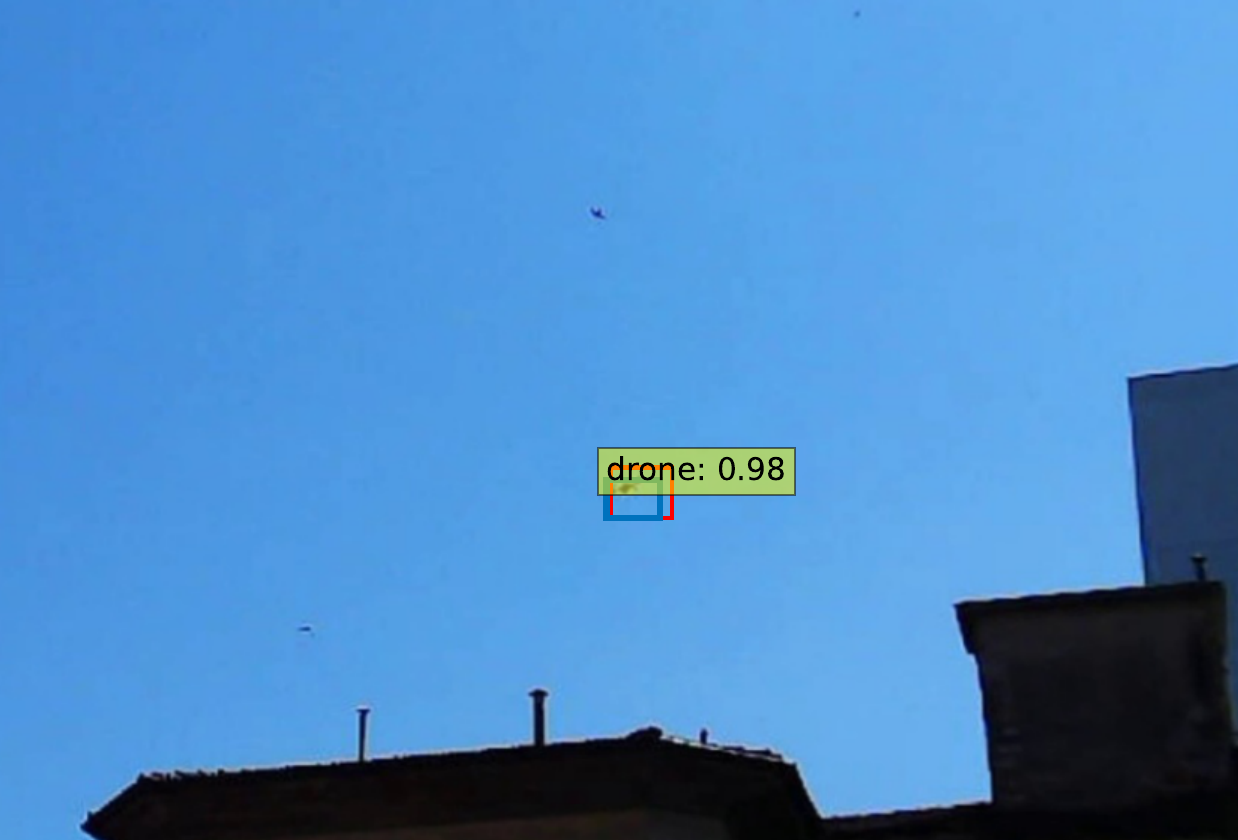} & \includegraphics[trim={250 50 50 50},clip,width=\figreswidth]{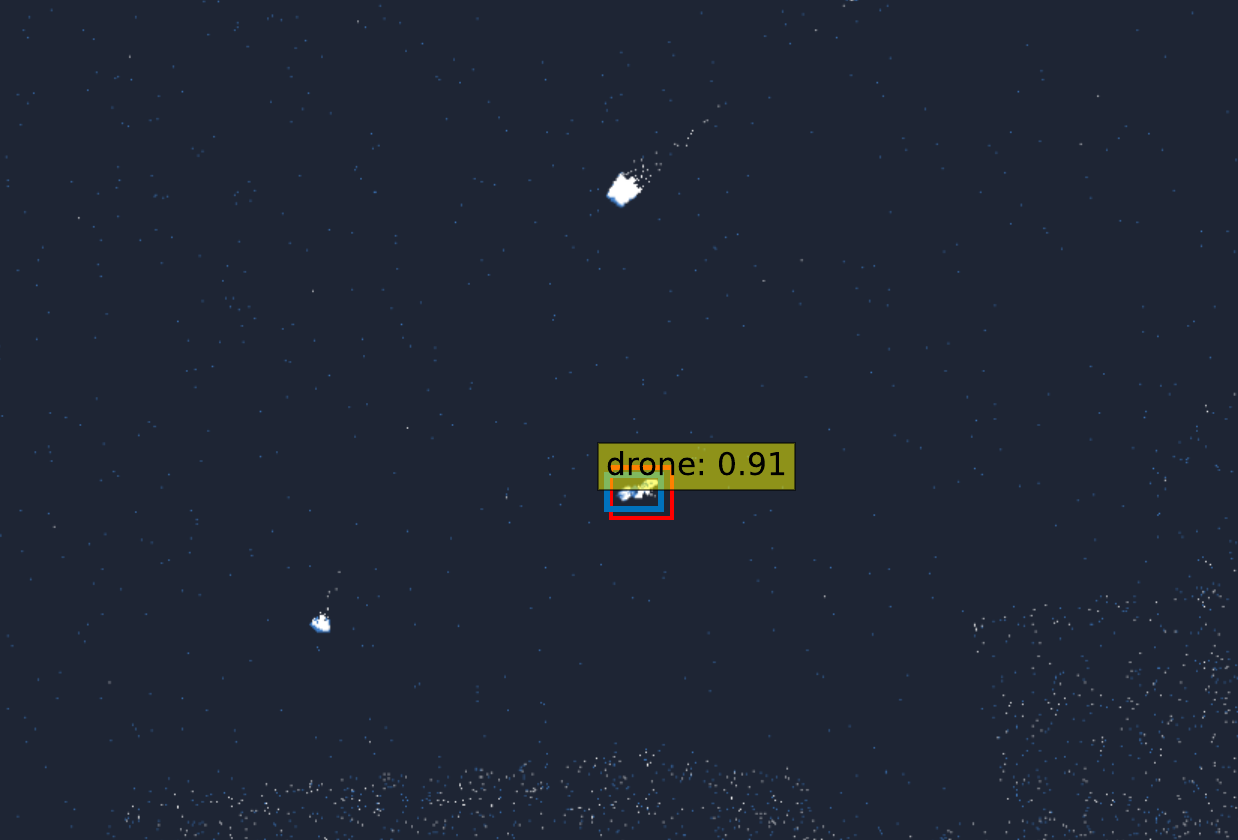}   & \includegraphics[trim={250 50 50 50},clip,width=\figreswidth]{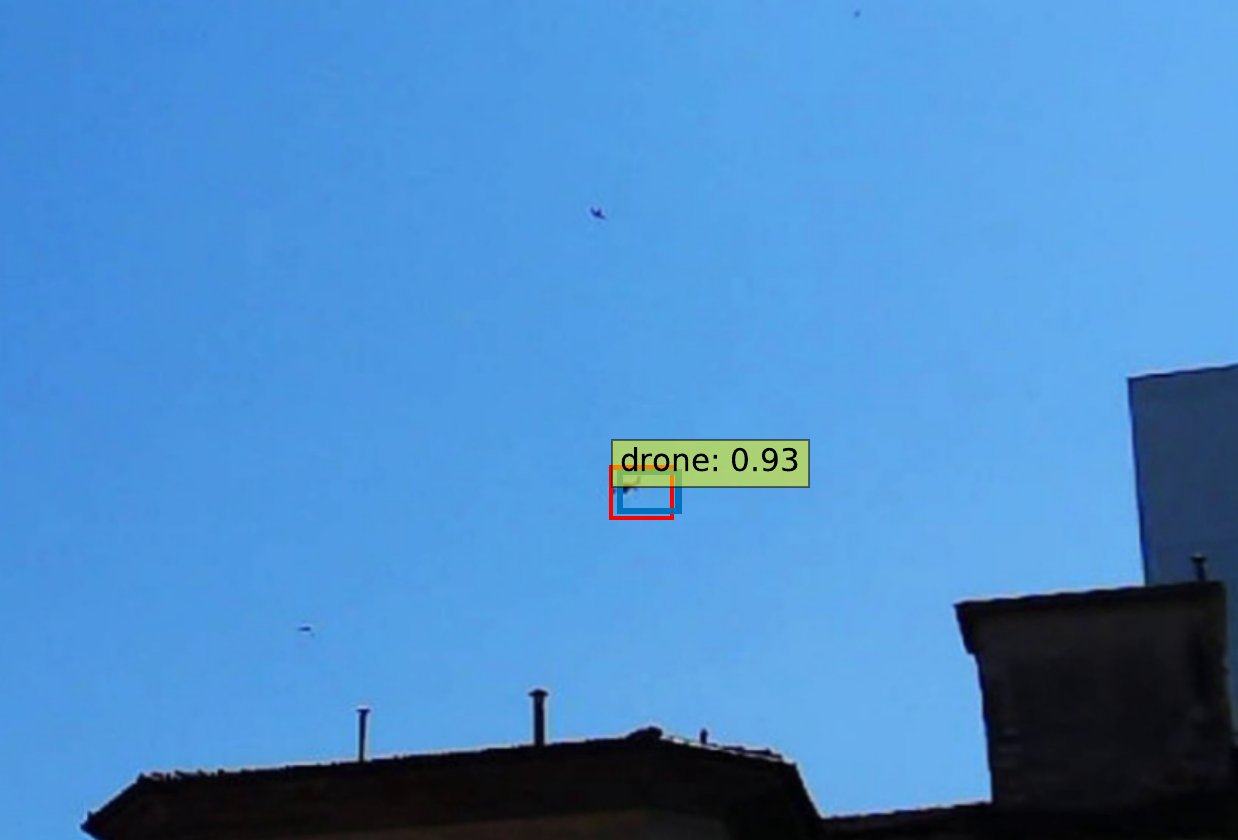}  & \includegraphics[trim={250 50 50 50},clip,width=\figreswidth]{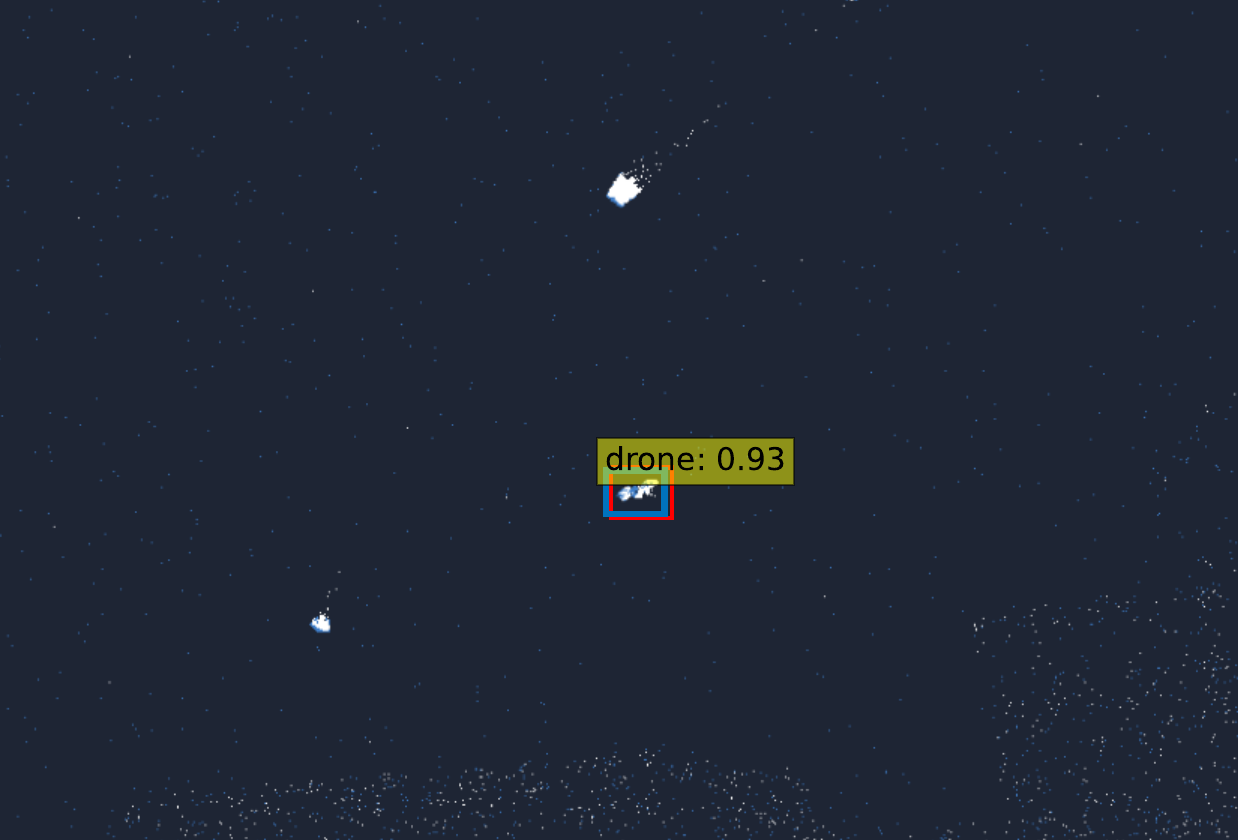}  & \includegraphics[trim={250 50 50 50},clip,width=\figreswidth]{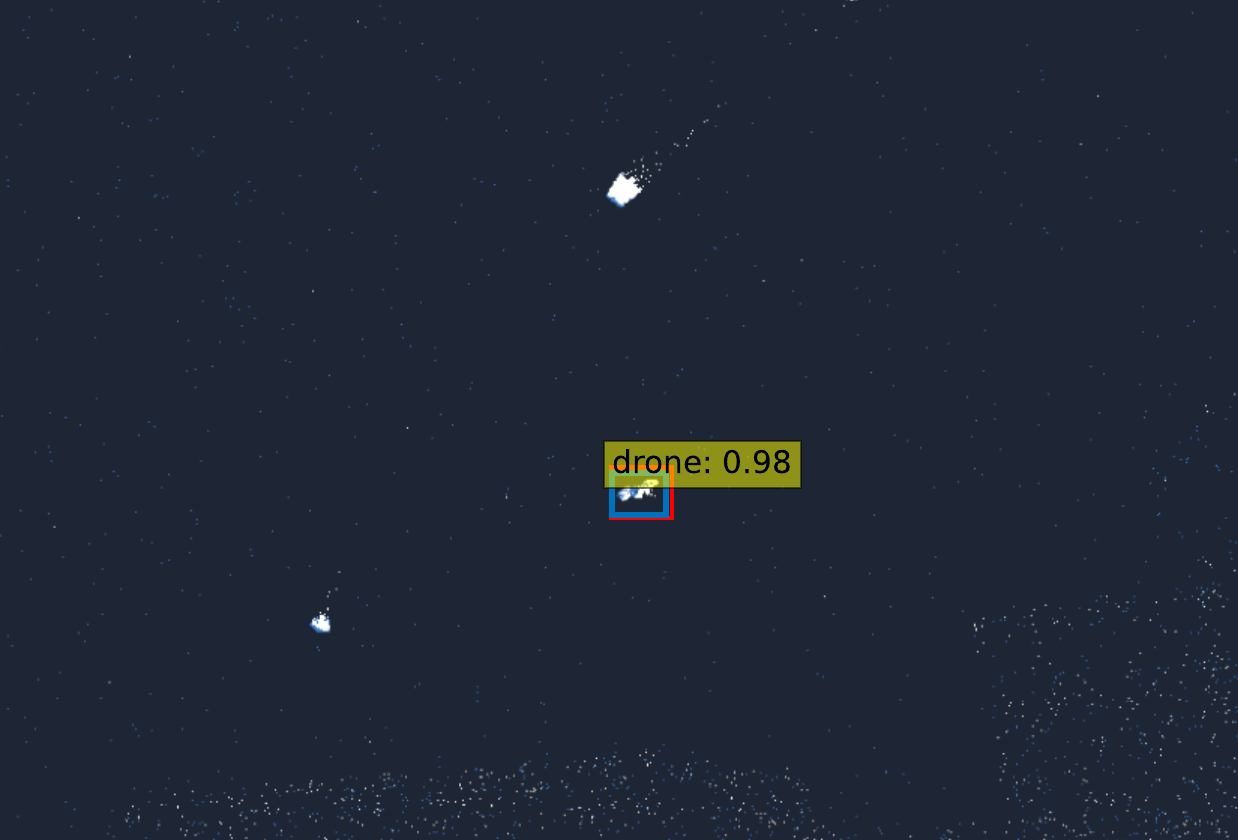}\\
\end{tabular}
\end{center}
\caption{Results of the main methods in various scenarios. The ground truth bounding box is shown in red, while the other colored boxes are the outputs of the model. EV-to-RGB and RGB-to-EV represent the results of the asymmetric modality injection fusion strategy. When only the red ground truth box is displayed, the model fails to detect the drone. Crops of the original images are displayed for better visualization.}
\label{fig:detres}
\end{figure}

\setlength{\tabcolsep}{2pt}
\begin{table}[t]
\begin{center}
\caption{Results of the different proposed architectures.}
\label{table:results}
\begin{tabular}{l|cccc}
\textbf{Model} & ~\textbf{AP50}~ & ~\textbf{AP50:95}~ & ~\textbf{AP75}~ & ~\textbf{Parameters}~~ \\
\hline
DETR Event~\cite{Carion-2020} & 80.5 & 34.8 & 21.6 & 41.302.368 \\
DETR RGB~\cite{Carion-2020} & 32.7 & 9.1 & 2.0 & 41.302.368 \\
\hline
Asymmetric RGB-to-EV~~ & 84.4 & 39.0 & \textbf{27.4} & 60.746.247 \\
Asymmetric EV-to-RGB~~ & 40.8 & 13.0 & 3.8 & 60.746.247 \\
\hline
Symmetric Fusion & 80.9 & 33.6 & 18.7 & 90.869.255 \\
\hline
Pooling (Encoder) & \textbf{85.2} & \textbf{39.3} & 27.2 & 59.166.983 \\
\end{tabular}

\end{center}
\end{table}
\setlength{\tabcolsep}{1.4pt}

We present in Tab. \ref{table:results} the results in terms of Average Precision (AP) with different intersection over union thresholds, namely 0.5, 0.75 and averaging the thresholds from 0.5 to 0.95 with a 0.05 step, as commonly done in datasets like COCO \cite{lin2014microsoft}.
The gap between the event and the RGB base DETR models appears immediately clear, underlying the difficulty of detecting drones in RGB frames. On the contrary, event data proves to be very effective.
The asymmetric modality injection fusion strategies still exhibit such a gap. Here, we refer to the two variants of the models as \texttt{x-to-y}, where \texttt{x} is the main modality and \texttt{y} the complementary one. At the same time though, injecting information from the other domain helps in significantly improving the AP of both models compared to their single modality counterparts.
Quite surprisingly, we found that the symmetric fusion, which combines the two asymmetric injections in a single model, fails to gain any improvement. We attribute this behavior to the high number of trainable parameters, which makes the model harder to train and more prone to overfitting.
Yet, the best results in terms of AP50 and AP50:95 are obtained by the pooling fusion strategy, which averages features extracted after the encoder layer of the model. The model however is still comparable with the RGB-to-EV, which achieves the best result in the challenging AP75 metric. Comparative qualitative results are visible in Fig. \ref{fig:detres}.

\subsection{Ablation Study}
\label{sec:ablation}
We also carried out a set of ablation studies to get a better understanding of the differences between the architectures. 
In particular, we investigate both the impact of the number of queries and the effect of fusing RGB and event data in different parts of the architecture.

\custompar{Early vs Late Fusion}
We investigated the effect of different variants of the same fusion strategy, by applying it picking different cut-off layers after which to apply the fusion module.
We start by comparing our best performing model, based on pooling fusion after the encoder layer, against a similar approach, with an earlier fusion. In this case, we directly average the features that come out from the ResNet backbone and then feed the resulting features to the transformer block. The results are shown in Tab. \ref{table:ablation_pooling}. An early fusion in this case is detrimental to the performance, especially for AP75, which exhibits a drop of 10 points. Interestingly, the AP50 decreases by only 0.5 points, suggesting that the detector still keeps working, yet it becomes less precise in identifying the exact boundaries of the drones.
Similarly, in Tab. \ref{table:ablation_symmetric}, we change the fusion point for the symmetric fusion architecture. Instead of fusing the two modalities after the encoder layer, we tested a late-fusion approach, where the modalities are fused after the final decoder. In this case, the degradation is small but consistent across all metrics.
To summarize, using two different modalities brings considerable improvements, yet picking the correct layer where to perform the modality fusion can have a significant impact on the overall capacity of the model. It appears that picking an intermediate layer yields the best results. This does not come as a surprise, as it offers a compromise between the number of parameters to be trained and the number of layers that can benefit from a joint training, sharing information across modalities.


\begin{table}[t]
    \parbox{.49\linewidth}{
        \centering
        \caption{Pooling fusion at different stages of the model. We compare average pooling using features extracted either after the encoder or after the ResNet backbone.}
\label{table:ablation_pooling}
        \begin{tabular}{l|cccH}
\textbf{Pooling} & ~\textbf{AP50}~ & ~\textbf{AP50:95}~ & ~\textbf{AP75}~ & ~\textbf{Parameters} \\
\hline
Encoder & \textbf{85.2} & \textbf{39.3} & \textbf{27.2} & 59.166.983 \\
ResNet & 84.7 & 35.2 & 17.9 & 51.276.551 \\
\end{tabular}
        
    }
    \hfill ~
    \parbox{.49\linewidth}{
        \centering
        \caption{We apply the event-to-RGB and RGB-to-event asymmetric fusions, followed by an average pooling at different stages of the model.}
\label{table:ablation_symmetric}
        \begin{tabular}{l|cccH}
\textbf{Symm.} & ~\textbf{AP50}~ & ~\textbf{AP50:95}~ & ~\textbf{AP75}~ & ~\textbf{Num Parameters} \\
\hline
Encoder & \textbf{80.9} & \textbf{33.6} & \textbf{18.7} & 90.869.255 \\
Decoder & 79.9 & 33.4 & 17.9 & 87.710.727 \\
\end{tabular}
        
        }
\end{table}

\begin{table}[t]
\begin{center}
\caption{Comparison of performances for different numbers of queries in the Encoder-Fusion model.}
\label{table:queries}
\begin{tabular}{c|cccHHHHHH}
\textbf{Pooling (Encoder)} & \textbf{AP50} & \textbf{AP 50:95} &  \textbf{AP75} &\textbf{$P_{S}$} & \textbf{$P_{M}$} & \textbf{$P_{L}$} & \textbf{$R_{S}$} & \textbf{$R_{M}$} & \textbf{$R_{L}$} \\
\hline
\textbf{5 Queries} & \textbf{85.2} & \textbf{39.3} & 27.2 & 24.1 & 44.3 & 55.6 & 33.4 & 56.7 & 69.5 \\
\textbf{10 Queries}  & 82.2 & 38.3 & \textbf{27.7} & 24.0 & 42.8 & 50.7 & 29.9 & 56.6 & 68.8 \\
\textbf{25 Queries}  & 82.6 & 37.4 & 24.9 & 24.1 & 41.6 & 51.2 & 34.6 & 56.7 & 70.5 \\
\textbf{100 Queries} & 83.7 & 38.8 & 27.5 & 24.6 & 44.8 & 54.1 & 39.9 & 58.0 & 76.5 \\
\end{tabular}

\end{center}
\end{table}

\custompar{Number of Object Queries}
In the original DETR model, 100 object queries are passed to the final transformer decoder and put in relation to the image encoded features. The number of queries is crucial in the sense that defines the maximum number of detectable objects in a scene. In a more specific sense, it is also strictly linked to the Hungarian Matching loss, in which each query should learn to be uniquely assigned to a detection and vice-versa. Thus, we expect the query number to have an effect in the model performance and convergence. 
To assert this we finetuned the models on the original DETR model number of queries, i.e. 100, as well as 10 and 25 to investigate the middle ground. Importantly, the to-be-finetuned model also comes with a number of queries of 100: thus, the decoder queries weight should still be able to converge in a limited number of epochs.
We report in Tab. \ref{table:queries} the results for the different number of queries for our best-performing model, namely the model with the pooling fusion strategy from Tab. \ref{table:results}. Interestingly, increasing the number of queries does not improve the results, that gracefully degrade for 10 and 25 object queries, while slightly recovering on the 100 queries case. 

\section{Conclusion and Future Work}
\label{sec:conclusion}
In this paper, we studied the problem of detecting drones with an event camera. In particular, we focused on developing different modality fusion strategies, that can be summarized into three main categories: pooling-based fusion, asymmetric modality injection and symmetric fusion. We found that event-based models demonstrate large performance improvements compared to RGB counterparts, yet the two modalities combined can improve and bridge the limitations of both modalities. In order to carry out our experiments we also collected and presented NeRDD, a novel multimodal dataset comprising 3.5 hours of manually annotated and spatio-temporally synchronized event-RGB videos. We believe that publicly releasing the dataset will foster research in the field of neuromorphic object detection, in particular for drone detection. For future works, we plan to investigate how to leverage the temporal information contained in the event data for a more resilient detection, as well as explore more advanced applications such as tracking and forecasting.

\paragraph{\textbf{Acknowledgments}}

This research was, in part, funded by Leonardo S.p.A.

%
%
\bibliographystyle{splncs04}
\bibliography{event-object-detection}
\end{document}